\documentclass[10pt,twocolumn,letterpaper]{article}

\usepackage{iccv}
\usepackage{times}
\usepackage{epsfig}
\usepackage{graphicx}
\usepackage{amsmath}
\usepackage{amssymb}
\usepackage{multirow}
\usepackage{booktabs}
\usepackage{subfigure}
\usepackage{nicefrac}
\usepackage{wrapfig}
\usepackage{verbatim}

\usepackage{enumerate}
\usepackage[export]{adjustbox}

\newcommand{\lossfunc}[1]{\ell_{#1}}
\newcommand{\dgrad}[2]{\nicefrac{\partial{#1}}{\partial{#2}}}
\newcommand{\grad}[2]{\frac{\partial{#1}}{\partial{#2}}}
\newcommand{\pred}[1]{q_{#1}}
\newcommand{\anchor}{q_{*}}

\usepackage[breaklinks=true,bookmarks=false]{hyperref}

\iccvfinalcopy 

\def\iccvPaperID{3141} 

\ificcvfinal\fi
\begin{document}

\title{Anchor Loss: Modulating Loss Scale based on Prediction Difficulty}

\author{Serim Ryou\\
California Institute of Technology\\
\and
Seong-Gyun Jeong\\
CODE42.ai\\
\and
Pietro Perona\\
California Institute of Technology\\
}

\maketitle
\ificcvfinal\thispagestyle{empty}\fi

\begin{abstract}
We propose a novel loss function that dynamically re-scales the cross entropy based on prediction difficulty regarding a sample. Deep neural network architectures in image classification tasks struggle to disambiguate visually similar objects. Likewise, in human pose estimation symmetric body parts often confuse the network with assigning indiscriminative scores to them. This is due to the output prediction, in which only the highest confidence label is selected without taking into consideration a measure of uncertainty. In this work, we define the prediction difficulty as a relative property coming from the confidence score gap between positive and negative labels. More precisely, the proposed loss function penalizes the network to avoid the score of a false prediction being significant. To demonstrate the efficacy of our loss function, we evaluate it on two different domains: image classification and human pose estimation. We find improvements in both applications by achieving higher accuracy compared to the baseline methods.
\end{abstract}

\section{Introduction}

\begin{figure}[t]
    \hspace{-0.3cm}
    \includegraphics[width=\columnwidth]{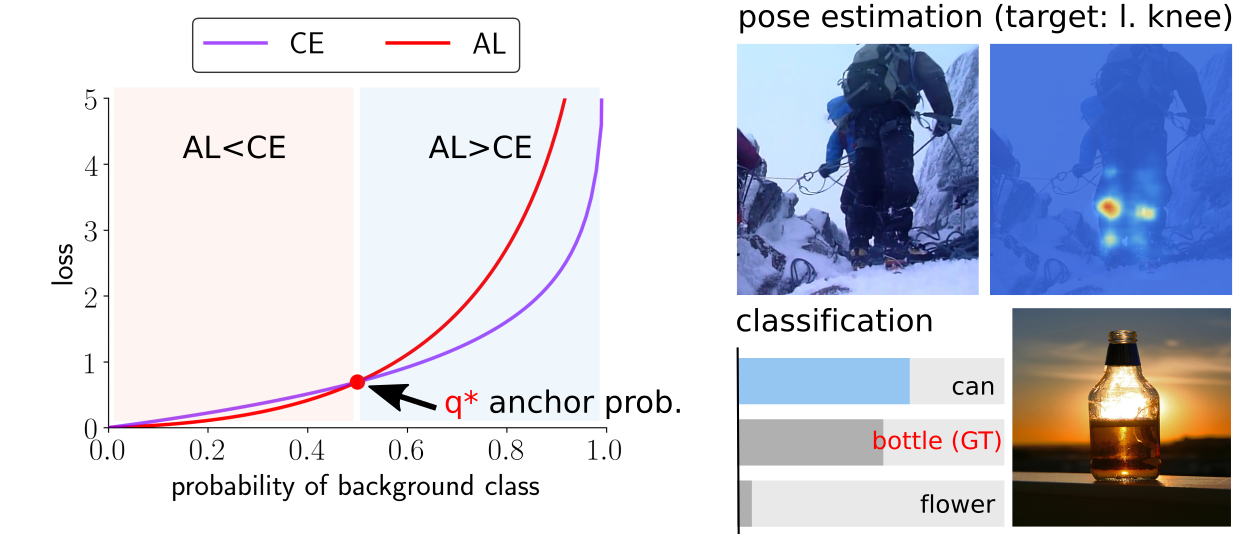}
    \caption{The overview of anchor loss. A network is confused about left-and-right body parts due to the symmetrical appearance of the human body, and struggles to disambiguate visually similar objects. Although the network output scores on the correct labels are relatively high, the final  prediction is always chosen by the index of the highest score, resulting in a wrong prediction. Our loss function is designed to resolve this issue by penalizing more than cross entropy when the non-target (background) probability is higher than the anchor probability. }\label{fig:motivation} 
\end{figure}

In many computer vision tasks, deep neural networks produce bi-modal prediction scores when the labeled sample point is confused with the other class. Figure~\ref{fig:motivation} illustrates some examples of network predictions with the presence of visually confusing cases. In all cases, though the network produces a non-trivial score about the correct label, the output prediction is wrong by taking the highest confidence label. For examples, human body parts are mostly composed of symmetric pairs. Even advanced deep architectures~\cite{He2016DeepRL,Newell2016StackedHN} are vulnerable to mistaking subtle differences of the left-and-right body parts~\cite{Ronchi_2017_ICCV}. Also, in image recognition, the output label confusion of look-alike instances is an unsolved problem~\cite{Hoiem_ECCV12}. Nevertheless, these tasks employ straightforward loss functions to optimize model parameters, \eg, mean squared error or cross entropy. 

In practice, look-alike instances incur an ambiguity in prediction scores, but it is hard to capture subtle differences in the network outputs by measuring the divergence of true and predicted distributions. Most classification tasks afterward make a final decision by choosing a label with the highest confidence score. We see that the relative score from the output distribution becomes an informative cue to resolve the confusion regarding the final prediction. We thus propose a novel loss function, which self-regulates its scale based on the relative difficulty of the prediction.

\begin{figure*}[t!]
\centering
	\subfigure[$\anchor=0.1$]{\includegraphics[width=5.8cm]{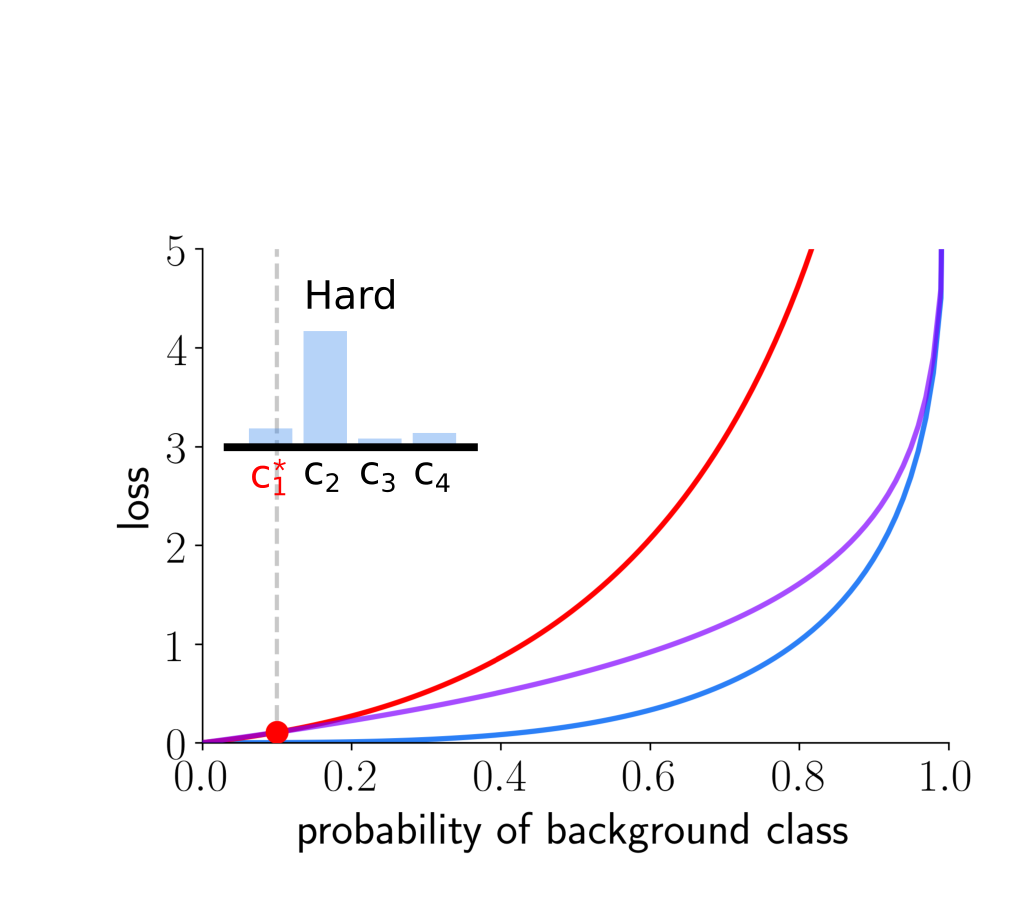}}\!
	\subfigure[$\anchor=0.5$]{\includegraphics[width=5.8cm]{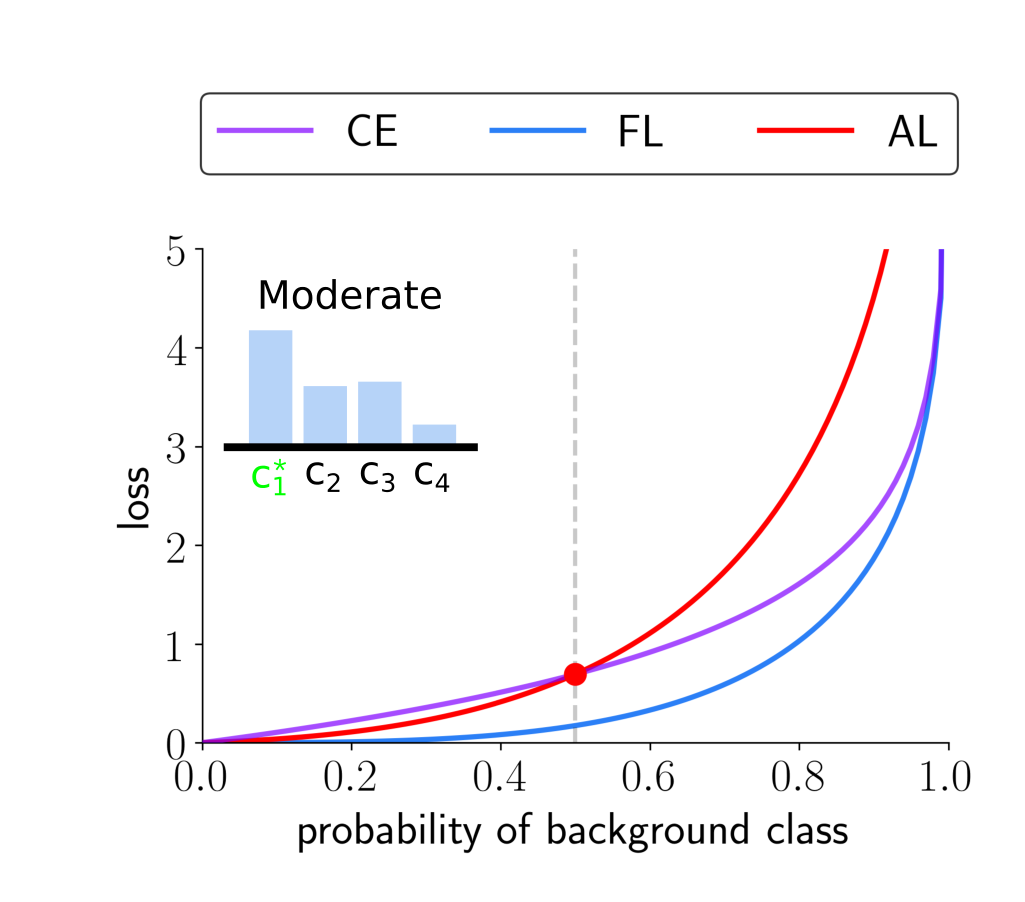}}\!
	\subfigure[$\anchor=0.9$]{\includegraphics[width=5.8cm]{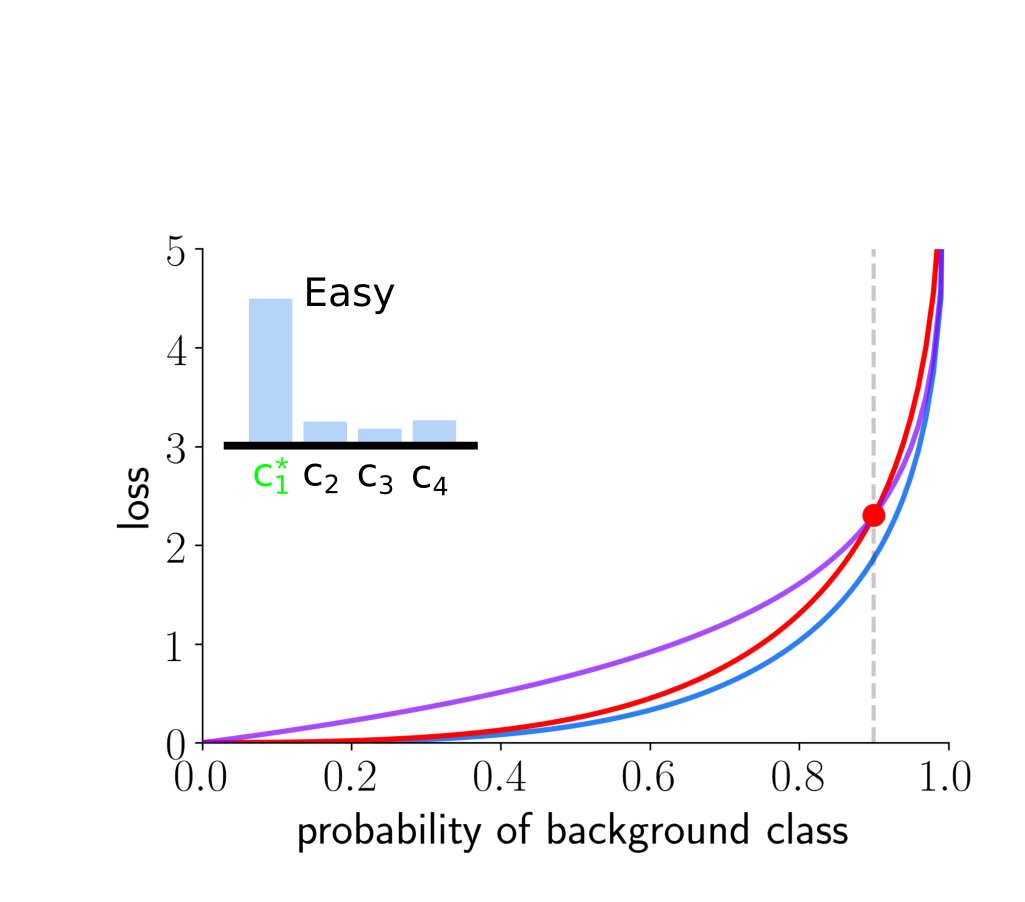}}\!
\caption{We depict how the anchor probability $\anchor$ affects our loss function compared to standard cross entropy (CE) and focal loss (FL)~\cite{FocalLoss17}. While FL always depresses the loss values for the samples producing trivial outcomes, anchor loss dynamically re-scales its loss values based on the relative difficulties of the target and the anchor probability. For these plots, the anchor probability is chosen as the prediction score ($\anchor = \pred{{C_1}}$) on the true positive label ($C_1$). Thus, if the networks produce higher score on the background label compared to the anchor, our loss encourages the network to correct the relative order of the predictions by penalizing more than the cross entropy. }\label{fig:anchors} 
\label{fig:plot}
\end{figure*}

We introduce {\em anchor loss} that adaptively reshapes the loss values using the network outputs. Specifically, the proposed loss function evaluates the prediction difficulties using the relative confidence gap between the target and background output scores, produced by the network, to capture the uncertainty. In other words, we increase the loss for hard samples (Figure~\ref{fig:anchors}a), while we down-weight the loss when a sample leads the network to assign a relatively high confidence score about the target class (Figure~\ref{fig:anchors}c). Finally, the anchor loss alleviates the need for a post-processing step by taking the prediction difficulty into account while training.

This idea, adjusting the loss scales based on prediction difficulty, has been applied to the task of object detection, which inherently suffers from severe class imbalance issue (countless background vs. scarce object proposals). Focal loss~\cite{FocalLoss17} is designed to overcome such class imbalance by avoiding major gradient updates on trivial predictions. However, while the focal loss uniformly down-weights easy samples to ignore, the proposed loss function leverages the confidence gap between the target and non-target output values to modulate the loss scale of the samples in the training phase. We define the prediction difficulty using a reference value which we call {\em anchor probability} $\anchor$ obtained from the network predictions. The way to pick an anchor probability becomes a design choice. One way to use it is by taking the target prediction score as an anchor probability to modulate the background (non-target) loss values. As depicted in Figure~\ref{fig:anchors}, the proposed loss function varies based on the anchor probabilities $\anchor$.  

We propose anchor loss for improving the prediction of networks on the most semantically confusing cases at training time. Specifically, the proposed anchor loss dynamically controls its magnitude based on prediction difficulty, defined from the network outputs. We observe that our loss function encourages the separation gap between the true labeled score and the most competitive hypothesis. Our main contributions are: (i) the formulation of a novel loss function (anchor loss) for the task of image classification (Section~\ref{ssec:loss_classif}); (ii) the adaptation of this loss function to human pose estimation (Section~\ref{ssec:loss_pose}); and (iii) a graphical interpretation about the behavior of the anchor loss function compared to other losses (Figure~\ref{fig:anchors} and \ref{fig:grad}). With extensive experiments, we show consistent improvements using anchor loss in terms of accuracy for image classification and human pose estimation tasks.

\section{Related Work}

\paragraph*{Class Imbalance Issue. }
Image classification task suffers class imbalance issue from the long-tail distribution of real-world image datasets. Typical strategies to mitigate this issue are class re-sampling~\cite{Chawla02smote,Han05,NN_BUDA2018} or cost-sensitive learning~\cite{Zhou_KDE06,CVPR_HuangLLT16,ICCV17_Dong}. Class re-sampling methods~\cite{Chawla02smote,NN_BUDA2018} redistribute the training data by oversampling the minority class or undersampling the majority class data. Cost-sensitive learning~\cite{CVPR_HuangLLT16,ICCV17_Dong} adjusts the loss value by assigning more weights on the misclassified minority classes. Above mentioned prior methods mainly focus on compensating scarce data by innate statistics of the dataset. On the other hand, our loss function renders prediction difficulties from network outputs without requiring prior knowledge about the data distributions.

\paragraph*{Relative Property in Prediction. }
Several researchers attempt to separate confidence scores of the foreground and background classes for the robustness~\cite{Gong14WARP,Zhang06_mll}. Pairwise ranking~\cite{Gong14WARP} has been successfully adopted in the multi-label image classification task, but efficient sampling becomes an issue when the vocabulary size increases. From the idea of employing a margin constraint between classes, L-softmax loss~\cite{LiuWYY16} combines the last fully-connected layer, softmax, and the cross entropy loss to encourage intra-class compactness and inter-class separability in the feature space. While we do not regularize the ordinality of the outputs, our loss function implicitly embodies the concept of ranking. In other words, the proposed loss function rules out a reversed prediction about target and background classes with re-scaling loss values.

\paragraph*{Outliers Removal vs. Hard Negative Mining.} 
Studies about robust estimation~\cite{huber_1964,Zhang_ICML04}, try to reduce the contribution on model parameter optimization from anomaly samples. Specifically, noise-robust losses~\cite{NIPS2018_Hendrycks,NIPS2018_Zhang,ICML18_Ren} have been introduced to support the model training even in the presence of the noise in annotations. Berrada~\etal~\cite{Berrada18} address the label confusion problem in the image classification task, such as incorrect annotation or multiple categories present in a single image, and propose a smooth loss function for top-$k$ classification.  Deep regression approaches~\cite{BarronCVPR2019,Belagiannis15} reduce the impact of outliers by minimizing M-estimator with various robust penalties as a loss function. Barron~\cite{BarronCVPR2019} proposed a generalization of common robust loss functions with a single continuous-valued robustness parameter, where the loss function is interpreted as a probability distribution to adapt the robustness.

On the contrary, there have been many studies with an opposite view in various domains, by handling the loss contribution from hard examples as a significant learning signal. Hard negative mining, originally called $\textit{Bootstrapping}$~\cite{Sung96}, follows an iterative bootstrapping procedure by selecting background examples for which the detector triggers a false alarm. Online hard example mining (OHEM)~\cite{ShrivastavaGG16} successfully adopts this idea to train deep ConvNet detectors in the object detection task. Pose estimation community also explored re-distributing gradient update based on the sample difficulty. Online Hard Keypoint Mining (OHKM)~\cite{CPN17} re-weights the loss by sampling few keypoint heatmaps which have high loss contribution, and the gradient is propagated only through the selected heatmaps. Our work has a similar viewpoint to the latter works to put more emphasis on the hard examples.

\paragraph*{Focal Loss.} 
One-stage object detection task has an inherent class imbalance issue due to a huge gap between the number of proposals and the number of boxes containing real objects. To resolve this extreme class imbalance issue, some works perform sampling hard examples while training~\cite{ShrivastavaGG16,FelzenszwalbGM10,SSD15}, or design a loss function~\cite{FocalLoss17} to reshape loss by down-weighting the easy examples. Focal loss~\cite{FocalLoss17} also addresses the importance of learning signal from hard examples in the one-stage object detection task. Without sampling processes, focal loss efficiently rescales the loss function and prevents the gradient update from being overwhelmed by the easy-negatives. Our work is motivated by the mathematical formulation of focal loss~\cite{FocalLoss17}, where predefined modulating term increases the importance of correcting hard examples. 

\paragraph*{Human Pose Estimation.}
Human pose estimation is a problem of localizing human body part locations in an input image. Most of the current works~\cite{Newell2016StackedHN,CPN17,Wei_CVPR16,YangLOLW17,KeECCV18,Tang_2018_ECCV} use a deep convolutional neural network and generate the output as a 2D heatmap, which is encoded as a gaussian map centered at each body part location. Hourglass network~\cite{Newell2016StackedHN} exploits the iterative refinements on the predictions from the repeated encoder-decoder architecture design to capture complex spatial relationships. Even with deep architectures, disambiguating look-alike body parts remain as a main problem~\cite{Ronchi_2017_ICCV} in pose estimation community. Recent methods~\cite{YangLOLW17,ChuYOMYW17,KeECCV18}, built on top of the hourglass network, use multi-scale and body part structure information to improve the performance by adding more architectural components. 

While there has been much interest in finding a good architecture tailored to the pose estimation problem, the vast majority of papers simply use mean squared error (MSE), which computes the L2 distance between the output and the prediction heatmap, as a loss function for this task. OHKM~\cite{CPN17}, which updates the gradient from the selected set of keypoint heatmaps, improves the performance when properly used in the refinement step. On the other hand, we propose a loss scaling scheme that efficiently redistributes the loss values without sampling hard examples. 

\section{Method}
In this section, we introduce {\em anchor loss} and explain the design choices for image classification and pose estimation tasks. First, we define the prediction difficulty and provide related examples. We then present the generalized form of the anchor loss function. We tailor our loss function on visual understanding tasks: image classification and human pose estimation. Finally, we give theoretical insight in comparison to other loss functions.

\subsection{Anchor Loss}\label{ssec:definition}

The inference step for most classification tasks chooses the label index corresponding to the highest probability. 
Figure~\ref{fig:motivation} shows sample outputs from the model trained with cross entropy. Although optimizing the networks with the cross entropy encourages the predicted distribution to resemble the true distribution, it does not convey the relative property between the predictions on each class.

Anchor loss function dynamically reweighs the loss value with respect to prediction difficulty. The prediction difficulty is determined by measuring the divergence between the probabilities of the true and false predictions. Here the anchor probability $\anchor$ becomes a reference value for determining the prediction difficulty. The definition of anchor probability $\anchor$ is arbitrary and becomes a design choice. However, in practice, we observed that setting anchor probability to the target class prediction score gives the best performance, so we use it for the rest of the paper. With consideration of the prediction difficulties, we formulate the loss function as follows:
\begin{align}
    \lossfunc{}(p,\pred{};\gamma) = -\underbrace{(1+\overbrace{\pred{}-\anchor}^{\textrm{prediction difficulty}})^\gamma}_{\textrm{modulator}}\underbrace{(1-p)\log(1-\pred{})}_{\textrm{cross entropy}},
\end{align}
where $p$ and $\pred{}$ denote empirical label and predicted probabilities, respectively. 
The anchor probability $\anchor$ is determined by the primitive logits, where the anchor is the prediction score on the true positive label.
Here, $\gamma \geq 0$ is a hyperparameter that controls the dynamic range of the loss function. 
Our loss is separable into two parts: modulator and cross entropy. The modulator is a monotonic increasing function that takes relative prediction difficulties into account, where the domain is bounded by $|\pred{}-\anchor|<1$. Suppose $\anchor$ be the target class prediction score. In an easy prediction scenario, the network assigns a correct label for the given sample point; hence $\anchor$ will be larger than any $\pred{}$. We illustrate the prediction difficulties as follows:
\begin{itemize}
    \setlength\itemsep{0.1em}
    \item {\bf Easy case} ($\pred{} < \anchor$): the loss function is suppressed, and thus rules out less informative samples when updating the model; 
    \item {\bf Moderate case} ($\pred{} = \anchor$): the loss function is equivalent to cross entropy, since the modulator becomes $1$; and
    \item {\bf Hard case} ($\pred{} > \anchor$): the loss function penalizes more than cross entropy for most of the range, since the true positive probability $\anchor$ is low.
\end{itemize}
As a result, we apply different loss functions for each sample.


\subsection{Classification}\label{ssec:loss_classif}
For image classification, we adopt sigmoid-binary cross entropy as a basic setup to diversify the way of scaling loss values. Unlike softmax, sigmoid activation handles each class output probability as an independent variable, where each label represents whether the image contains an object of corresponding class or not. 
This formulation also enables our loss function to capture subtle differences from the output space by modulating the loss values on each label. 

For image classification, we obtained the best performance when we set the anchor probability to the output score of the target class. 
The mathematical formulation becomes as follows:


\begin{align}
&\lossfunc{cls}(p,\pred{};\gamma) \\\nonumber
&= - \sum_{k=1}^{K} p_k \log \pred{k} + (1 - p_k)(1 + \pred{k} - \pred{*})^\gamma \log (1 - \pred{k}),
\end{align}

\noindent where $p_k$ and $q_k$ represent the empirical label and the predicted probability for class $k$. We add a margin variable $\delta$ to anchor probability $\pred{*}$ to penalize the output variables which have lower but close to the true positive prediction score. Thus the final anchor probability becomes $\pred{*} = q_t - \delta$, where $t$ represents the target index ($p_t=1$), and we set $\delta$ to $0.05$.
\subsection{Pose Estimation}\label{ssec:loss_pose}

\begin{figure}[t]
	\centering
    \subfigure[input]{\includegraphics[width=2.5cm]{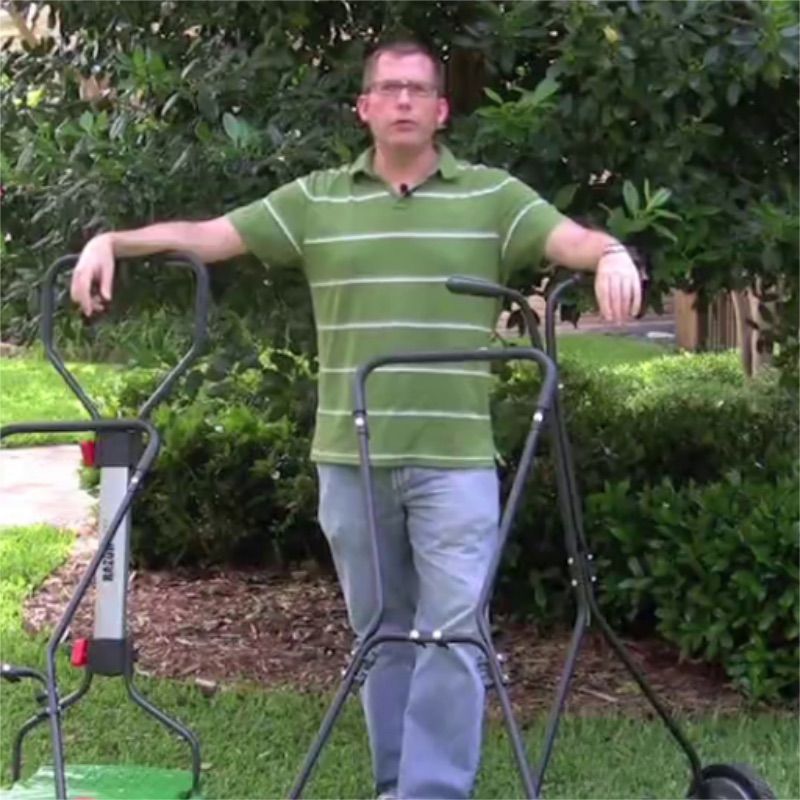}}
    \subfigure[heatmap]{\includegraphics[width=2.5cm]{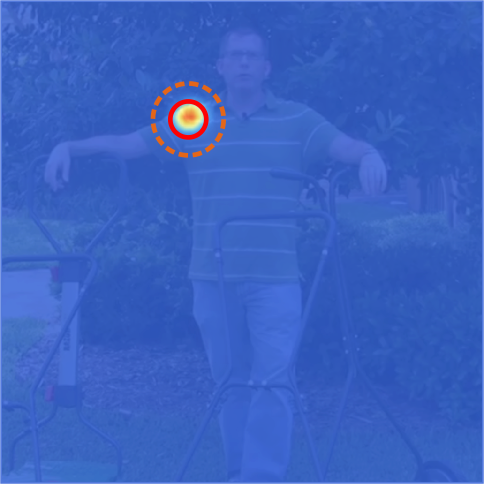}}
    \subfigure[mask]{\includegraphics[width=2.5cm]{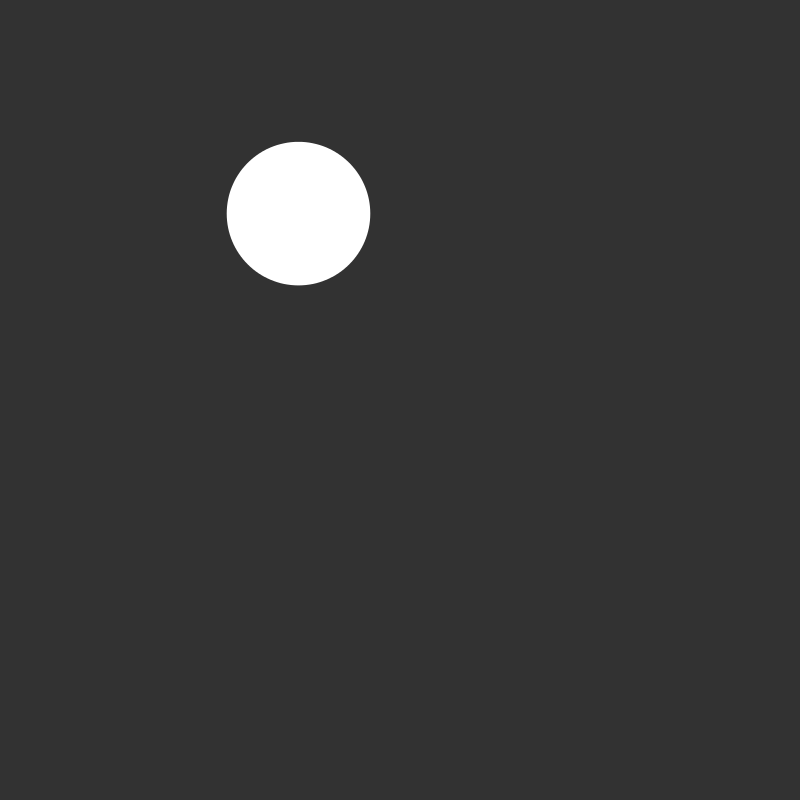}}
    \caption{How an anchor probability is chosen for the pose estimation task. For the target body part of right shoulder (b), the maximum confidence score inside the solid red circle becomes an anchor probability to modulate the loss values in mask areas (c).}\label{fig:pose_anchor}
\end{figure}

Current pose estimation methods generate a keypoint heatmap for each body part at the end of the prediction stage, and predict the pixel location that has the highest probability. The main difference of pose estimation and object classification tasks is that the target has spatial dependency between adjacent pixel locations. As a result, assigning a single pixel as the true positive may incur a huge penalty on adjacent pixels. To alleviate this issue, we adopt a gaussian heatmap centered on the target keypoint as the same encoding scheme as the previous works~\cite{Newell2016StackedHN,Wei_CVPR16,CPN17}, and apply our loss function on only true negative pixels ($p_i = 0$). In other words, we use a mask variable $M(p)$ to designate the pixel locations where our loss function applies, and use standard binary cross entropy on unmasked locations. 

\begin{align}
    M(p) =\left\{
	\begin{array}{ll}
		1& \text{if $p = 0$,}\\
		0& \text{otherwise.}\\
	\end{array}
	\right.
\end{align}

As in object classification, we found that using true-positive probability value to penalize background pixel locations gives better performance. Considering the spatial dependency, anchor probabilities are chosen spatially from the circle of high confidence, where the ground truth probability is greater than 0.5. That is,
\begin{align}
    q_* = \max_{i \forall p_i > 0.5} q_i,
\end{align}

\noindent We illustrate this procedure in Figure~\ref{fig:pose_anchor}. For simplicity, we denote the standard binary cross entropy as $\lossfunc{BCE}$. Finally, our loss function for pose estimation problem is defined as:
\begin{align}
    \lossfunc{pose}(p,\pred{};\gamma) = & \large[ M(p) * (1 + \pred{} - \pred{*})^\gamma \\\nonumber
     &+ (1 - M(p)) \large]* \lossfunc{BCE}(p,\pred{} ),
\end{align} 


%

\subsection{Relationship to Other Loss Functions}\label{ssec:review_losses}
Our goal is to design a loss function which takes the relative property of the inference step into account. In this section, we discuss how binary cross entropy (6) and focal loss~\cite{FocalLoss17} (7) relate to anchor loss. Let $p\in\{0,1\}$ denote the ground truth, and $q\in[0,1]$ represent predicted distribution. The loss functions are

\begin{align}
	\lossfunc{CE}(p, \pred{}) &= -\big[p\log(\pred{}) + (1-p)\log(1-\pred{})\big],\\
	\lossfunc{FL}(p, \pred{} ; \gamma) &= -\big[p(1-\pred{})^\gamma\log(\pred{}) + (1-p)\pred{}^\gamma\log(1-\pred{})\big],
\end{align}
For the sake of conciseness, we define the probability of ground truth as $\pred{t} = p\pred{} + (1-p)(1-\pred{})$. Then we replace the loss functions as follows:

\begin{align}
	\lossfunc{CE}(\pred{t}) &= - \log (\pred{t}),\\
	\lossfunc{FL}(\pred{t} ; \gamma) &= - (1-\pred{t})^\gamma \log (\pred{t}),
\end{align}

\noindent where $\pred{}$ represents the output vector from the network. The modulating factor $(1-\pred{t})^\gamma$ with focusing parameter $\gamma$ reshapes the loss function to down-weight easy samples. Focal loss was introduced to resolve the extreme class imbalance issue in object detection, where the majority of the loss is comprised of easily classified background examples. Object detection requires the absolute threshold value to decide the candidate box is foreground or background. On the other hand, classification requires the confidence score of the ground truth label to be higher than all other label scores.

If we set $\pred{*}=1-p$, which means $\pred{*}=1$ for the background classes and $\pred{*}=0$ for the target class:
\begin{align}
    \pred{*} = \left\{
	\begin{array}{ll}
		1& \text{$p = 0$ \quad background classes,}\\
		0& \text{$p = 1$ \quad target class,}\\
	\end{array}
	\right.
\end{align}

\noindent then the modulator becomes:
\begin{align}
    (1 - \pred{t} + \pred{*}) = \left\{
	\begin{array}{ll}
		(1 - (1 - \pred{}) + 1) = (1-\pred{}) & \text{$p = 0$,}\\
		(1 - \pred{} + 0) = \pred{} & \text{$p = 1$,}\\
	\end{array}
	\right.
\end{align}

\noindent and feeding this modulator value to anchor loss becomes a mathematical formulation of focal loss:
\begin{align}
    &\lossfunc{AL}(p,\pred{};\gamma) = - \big[ p ( 1 - \pred{})^\gamma \log (\pred{}) + (1 - p) \pred{}^\gamma \log (1 - \pred{}) \big],\nonumber\\
    &\text{where~}\pred{*}=1-p.
\end{align}
If we set $\gamma = 0$, the the modulator term becomes 1, and anchor loss becomes binary cross entropy.

\subsection{Gradient Analysis}\label{ssec:gradient}
\begin{figure}[t]
\begin{center}
   \subfigure[$\lossfunc{FL}(\pred{t};\gamma)$]{\includegraphics[width=4.15cm]{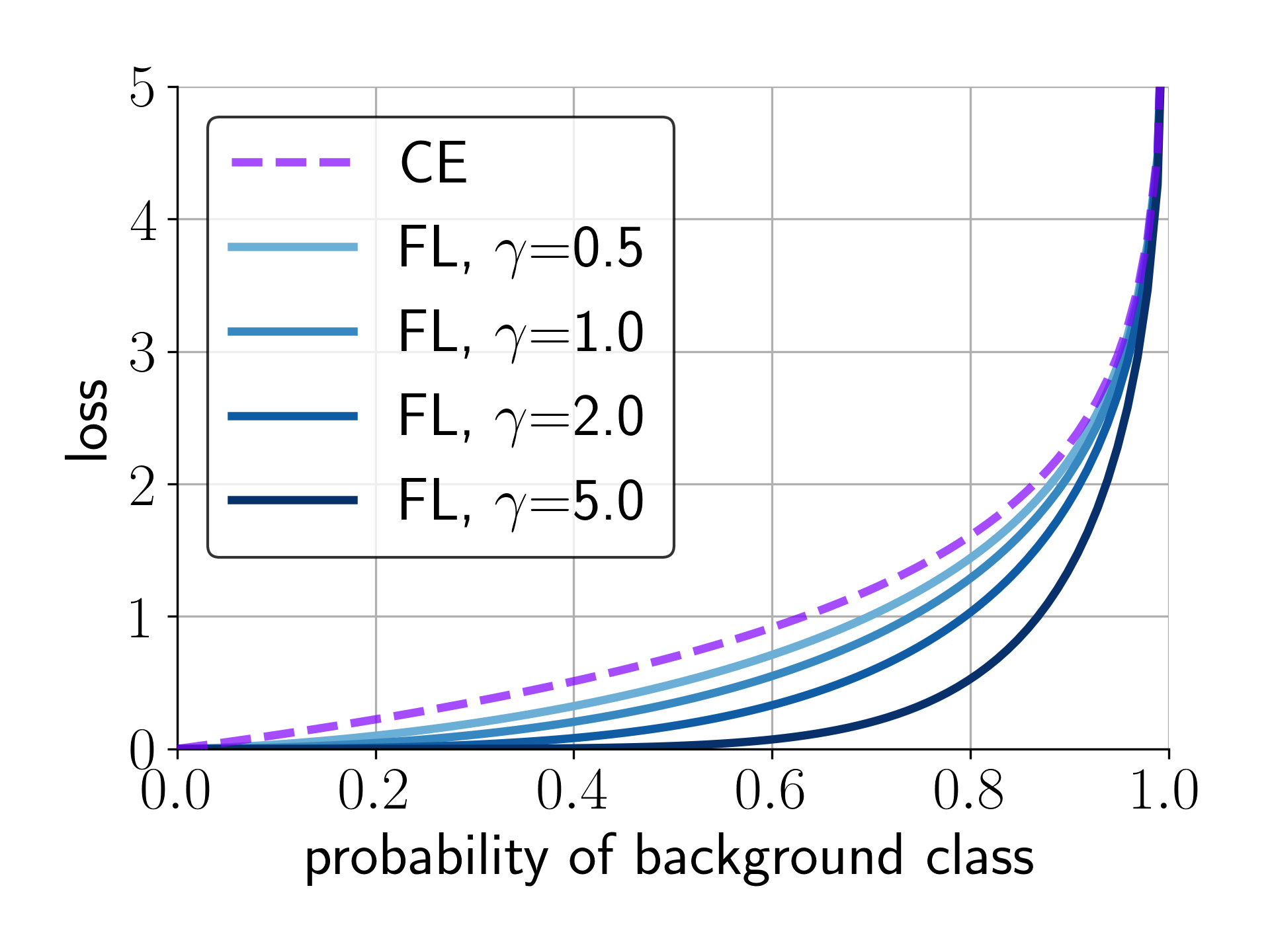}}\!
   \subfigure[$|\dgrad{\lossfunc{FL}}{\pred{t}}|$]{\includegraphics[width=4.15cm]{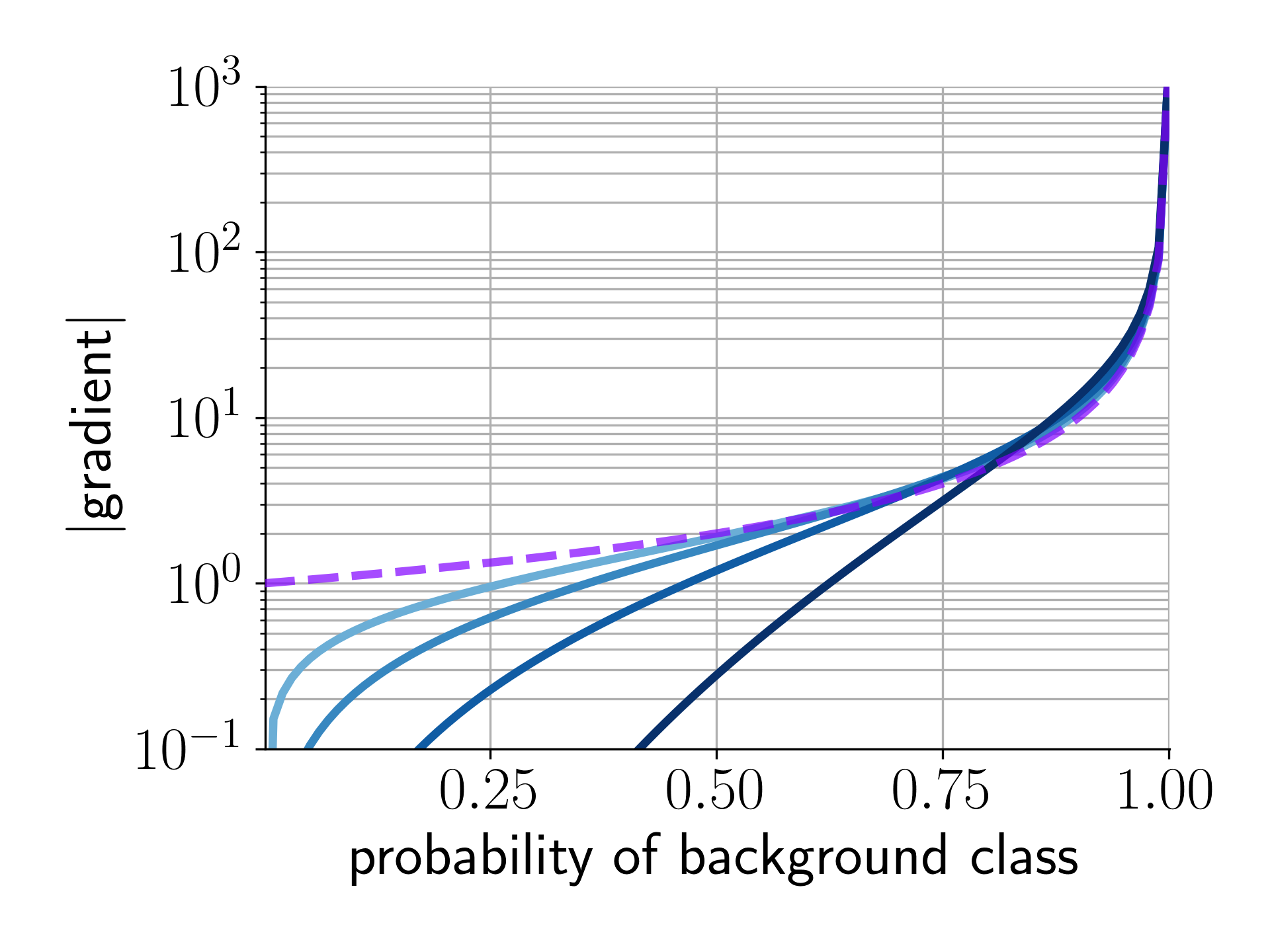}}\\
   \subfigure[$\lossfunc{AL}({\pred{t};\gamma}),\pred{*}=0.5$]{\includegraphics[width=4.15cm]{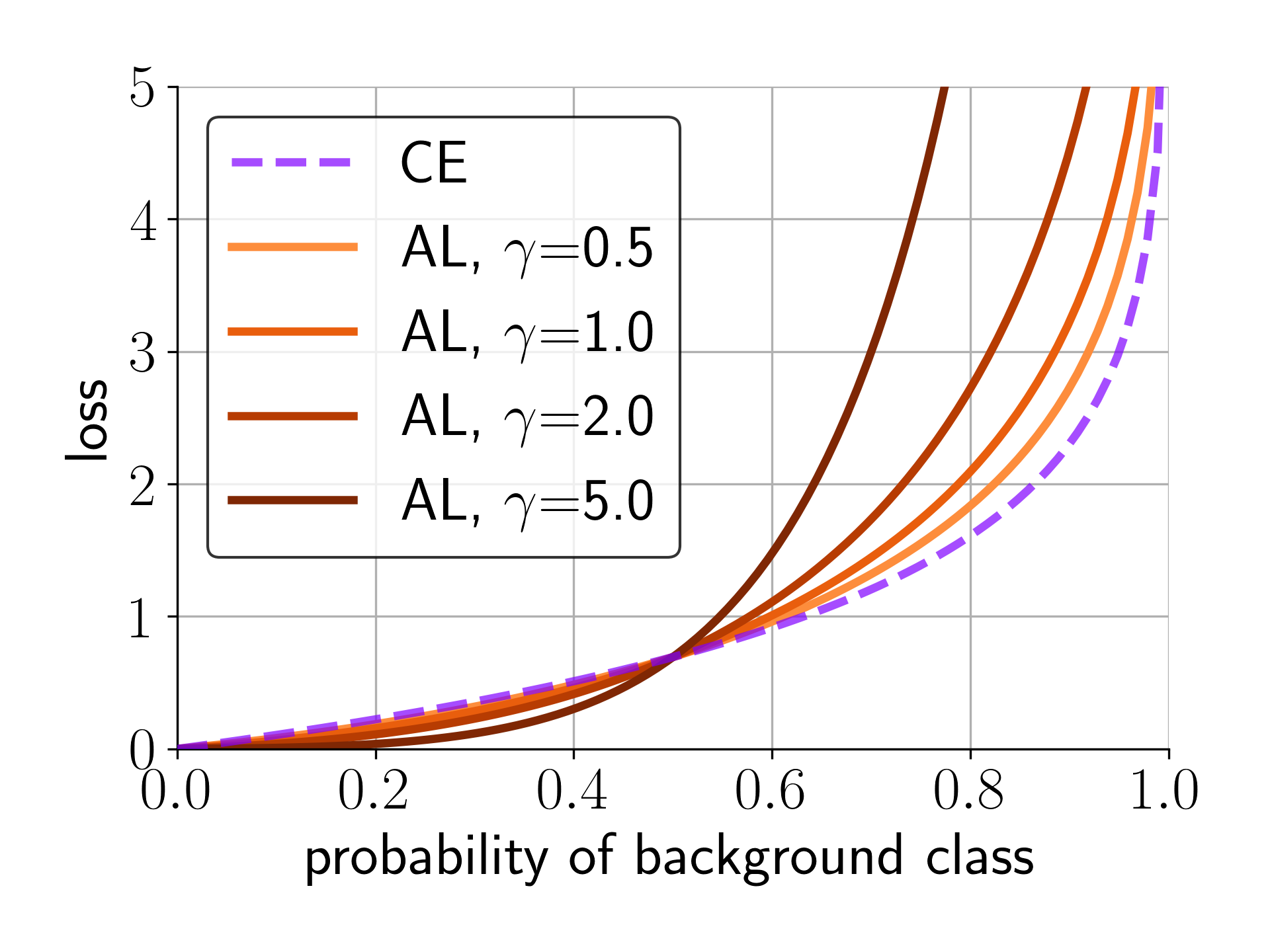}}\!
   \subfigure[$|\dgrad{\lossfunc{AL}}{\pred{t}}|$]{\includegraphics[width=4.15cm]{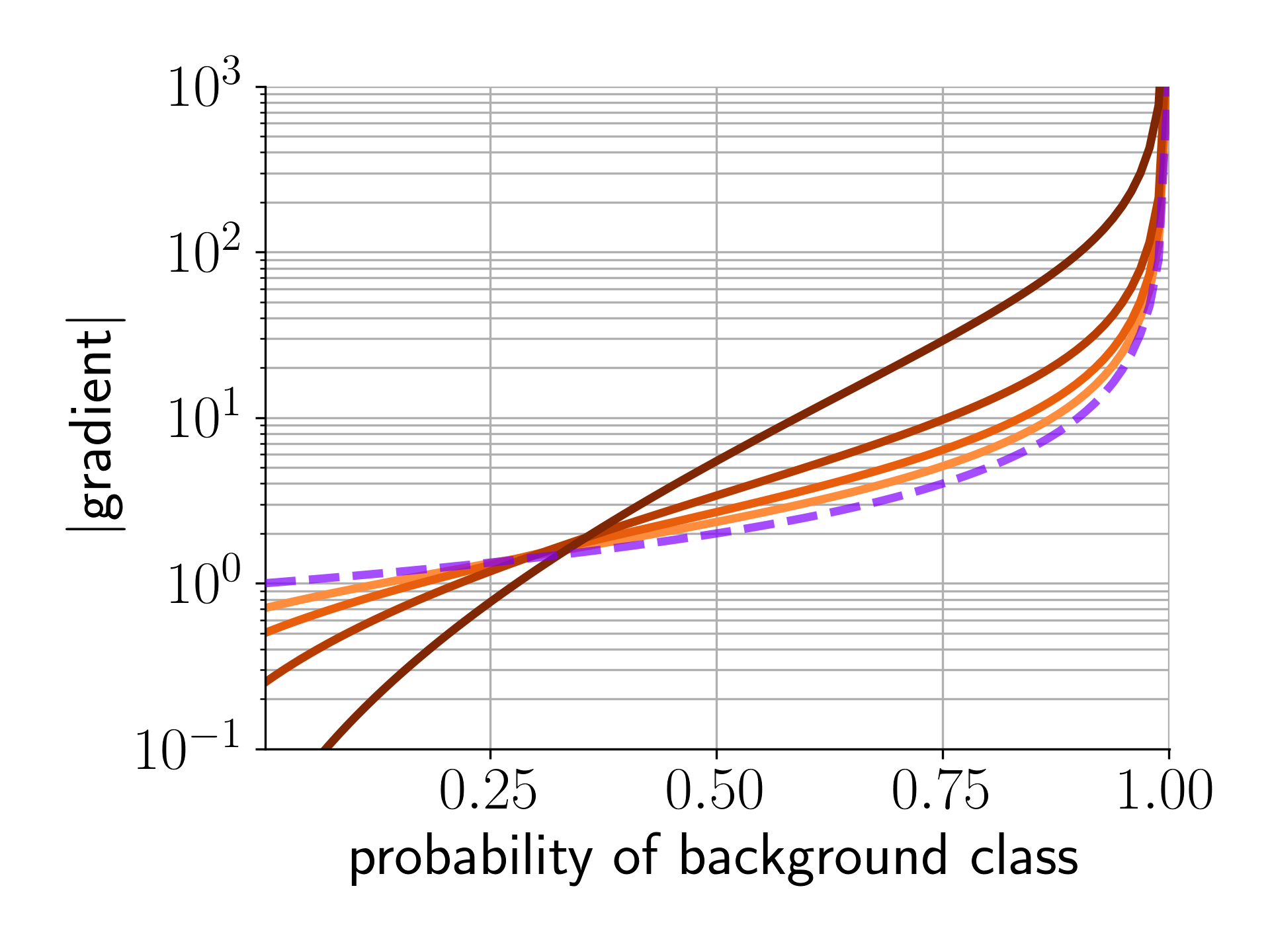}}
\end{center}
   \caption{Gradient figure: sample gradient output of background probability distribution. Compared to the cross entropy, the magnitude of gradient increases when the prediction is higher than the anchor probability.}
\label{fig:grad}
\end{figure}

We compute the gradient of our loss function and compare with the binary cross entropy and the focal loss. For simplicity, we focus on the loss of background label, which we discuss in Section~\ref{ssec:definition}. Note that we detach the anchor probability $\pred{*}$ while backpropagation and only use it as a scaling term in the modulator. 

\begin{align}
    \lossfunc{AL}(\pred{}) &= - (1 + \pred{} - \pred{*})^\gamma \log (1 - \pred{}) \\
    \grad{\lossfunc{AL}}{\pred{}}(\pred{}) &=
    - (1 + \pred{} - \pred{*})^{\gamma-1} \left[ \gamma \log (1-\pred{}) 
    - \frac{1+\pred{}-\pred{*}}{1-\pred{}}\right]
\end{align}


Figure~\ref{fig:grad} shows the gradient of our loss function, focal loss, and cross entropy. Compared to the cross entropy, the gradient values of focal loss are suppressed for all ranges. On the other hand, our loss function assigns larger gradient values when the prediction is higher than the anchor probability, and vice versa. 
\section{Experiments}

We conduct experiments on image classification and human pose estimation. In this section, we briefly overview the methods that we use in each domain, and discuss the experimental results. 

\subsection{Image Classification}

\paragraph*{Datasets.} For the object classification, we evaluate our method on CIFAR-10/100~\cite{KrizhevskyCIFAR} and ImageNet (ILSVRC 2012)~\cite{imagenet_cvpr09}. CIFAR 10 and 100 each consist of 60,000 images with 32$\times$32 size of 50,000 training and 10,000 testing images. In our experiment, we randomly select 5,000 images for the validation set. CIFAR-10 dataset has 10 labels with 6,000 images per class, and CIFAR-100 dataset has 100 classes each containing 600 images. 

\paragraph*{Implementation details.} For CIFAR, we train ResNet-110~\cite{He2016DeepRL} with our loss function and compare with other loss functions and OHEM. 
We randomly flip and crop the images padded with 4 pixels on each side for data augmentation. All the models are trained with PyTorch~\cite{paszke2017automatic}. Note that our loss is summed over class variables and averaged over batch. The learning rate is set to 0.1 initially, and dropped by a factor of 0.1 at 160 and 180 epochs respectively. In addition, we train ResNet-50 models on ImageNet using different loss functions. We use 8 GPUs and batch size of 224. To accelerate training, we employ a mixed-precision. We apply minimal data augmentation, \ie, random cropping of $224\times224$ and horizontal flipping. The learning rate starts from $0.1$ and decays $0.1$ every 30 epoch. We also perform learning rate warmup strategy for first 5 epochs as proposed in~\cite{He2016DeepRL}. 
%
\begin{table}
	\caption{Classification accuracy on CIFAR (ResNet-110)}\label{tab:cifar}
	\vspace{0.2cm}
	\centering
	\resizebox{0.45\textwidth}{!}{
	\begin{tabular}{l l c c c c c }
	\toprule
	 &&& \bf{CIFAR-10} &&\multicolumn{2}{c}{\bf{CIFAR-100}}\\
	 \cmidrule[0.5pt]{4-4}\cmidrule[0.5pt]{6-7}
	Loss Fn. & Parameter && Top-1 &&  Top-1 & Top-5 \\
	\midrule
	CE & && 93.91 $\pm$ 0.12 
	     && 72.98 $\pm$ 0.35 & 92.55 $\pm$ 0.30 \\ 
	BCE & && 93.69 $\pm$ 0.08 
	       && 73.88 $\pm$ 0.22 & 92.03 $\pm$ 0.42 \\ 
	OHEM & $\rho=0.9, 0.9$ && 93.90 $\pm$ 0.10 
	          && 73.03 $\pm$ 0.29 & \underline{92.61} $\pm$ 0.21 \\ 
	FL & $\gamma=2.0, 0.5$ && 94.05 $\pm$ 0.23 
	                             && 74.01 $\pm$ 0.04 & 92.47 $\pm$ 0.40 \\ \midrule
   \bf{Ours} &&&&&\\
	AL  & $\gamma=0.5, 0.5$  && \underline{94.10} $\pm$ 0.15 
                                     && \underline{74.25} $\pm$ 0.34 & \bf{92.62} $\pm$ 0.50 \\ 
   AL  w/ warmup & $\gamma=0.5, 2.0$  && \bf{94.17} $\pm$ 0.13 
                                     && \bf{74.38} $\pm$ 0.45 & 92.45 $\pm$ 0.05 \\                
	\bottomrule
	\end{tabular}}
\end{table}

\begin{table}
	\caption{Classification accuracies on ImageNet (ResNet-50)}\label{tab:imagenet}
	\vspace{0.2cm}
	\centering
    \scalebox{0.8}{%
	\begin{tabular}{l c c c }
	\toprule
	Loss Fn. & Parameter & Top-1 & Top-5 \\
	\midrule
	CE & & 76.39 & 93.20 \\
	
	OHEM & $\rho=0.8$ & 76.27	& 93.21\\
	FL & $\gamma=0.5$   & 76.72 & 93.06 \\
	AL (ours) & $\gamma=0.5$ & {\bf{76.82}} & 93.03 \\ 
	\bottomrule\\[-1em]
	\end{tabular}}
\end{table}

\begin{figure*}[t]
    \centering
    \begin{minipage}{.48\textwidth}
        \centering
        \includegraphics[width=8cm]{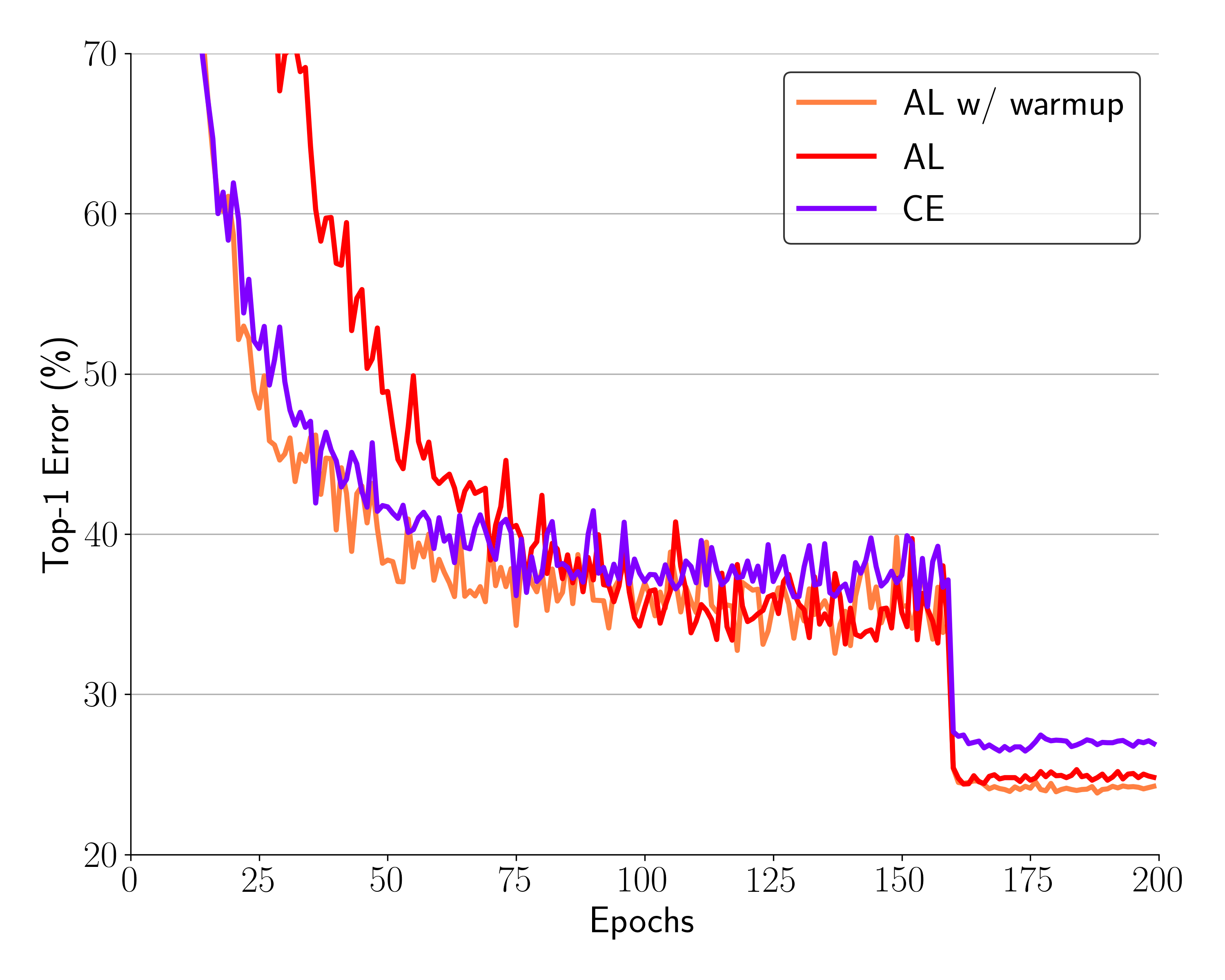}
        \caption{Validation curves of ResNet-110 on CIFAR-100 dataset. We compare our loss function to CE. }\label{fig:loss}
    \end{minipage}\hfill
    \begin{minipage}{.48\textwidth}
        \centering
        \includegraphics[width=8cm]{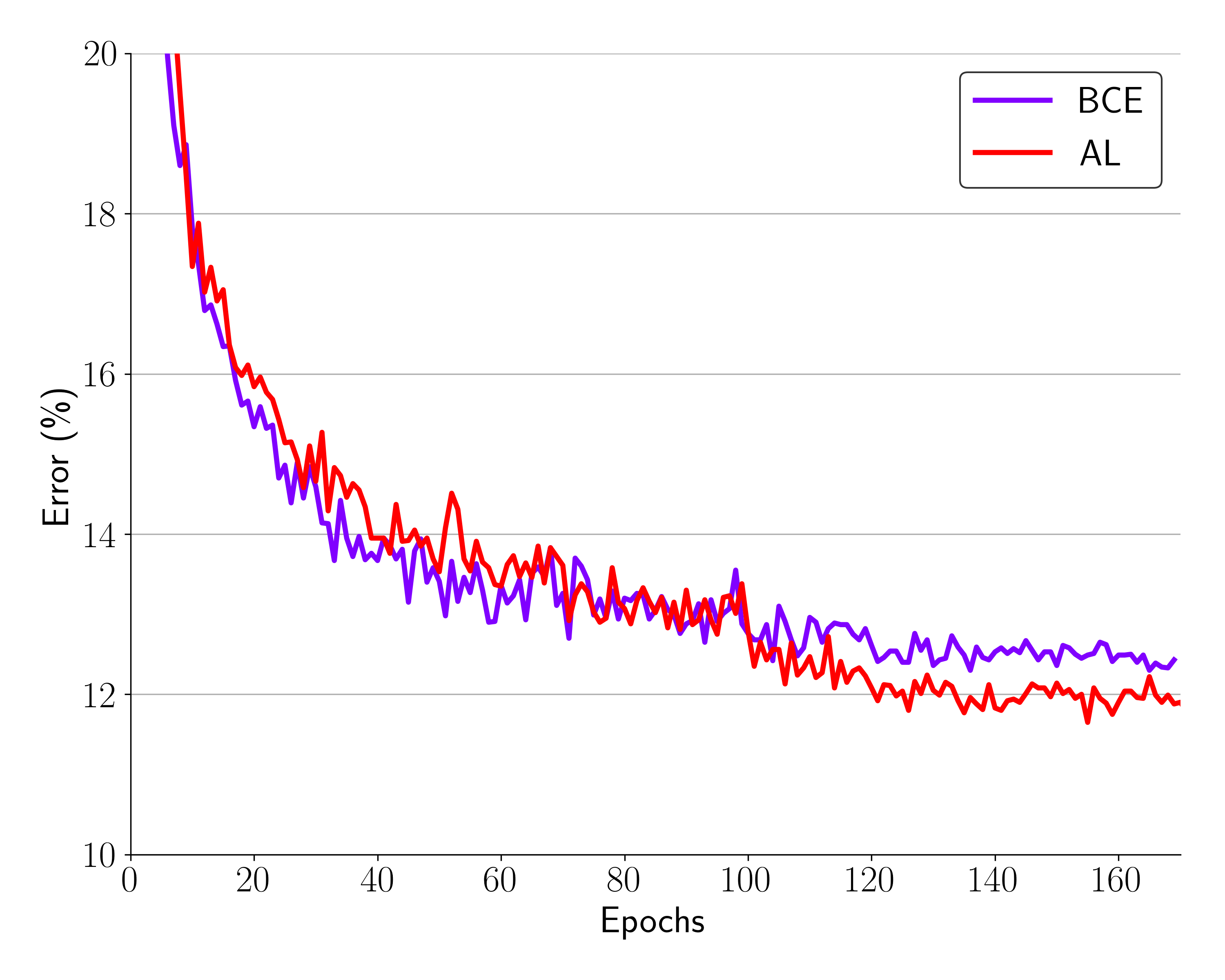}
        \caption{Validation curves of 2-stacked Hourglass on MPII dataset. We compare our loss function to BCE.}\label{fig:pose_loss}
    \end{minipage}
\end{figure*}


\paragraph*{Results.} For CIFAR, we train and test the network three times and report the mean and standard deviation in Table~\ref{tab:cifar}. We report top-1 and top-5 accuracy and compare the score with other loss functions and OHEM. OHEM computes the loss values for all samples in a batch, chooses the samples of high loss contribution with a ratio of $\rho$, and updates the gradient only using those samples. As we can see in the Table~\ref{tab:cifar}, our loss function has shown improvements over all loss functions we evaluated. 
For CIFAR 100, performance improved by simply replacing the cross entropy to the binary cross entropy, and anchor loss gives further gain by exploiting the automated re-scaling scheme. 
With our experimental setting, we found that sampling hard examples (OHEM) does not help. We tried out few different sampling ratio settings, but found performance degradation over all ratios.  

\paragraph*{Ablation Studies. } 
As an ablation study, we report the top-1 and top-5 accuracy on CIFAR-100 by varying the $\gamma$ in Table~\ref{tab:abs_cifar}. For classification task, low $\gamma$ yielded a good performance. We also perform experiments with fixed anchor probabilities to see how the automated sample difficulty from the network helps training. The results in Table~\ref{tab:abs_cifar} show that using the network output to define sample difficulty and rescale the loss based on this value helps the network keep a good learning signal.

\paragraph*{CE warmup strategy.}
To accelerate and stabilize the training process, we use CE for first few epochs and then replace loss function to AL. We tested CE warmup on CIFAR-100 for the first 5 epochs (Figure~\ref{fig:loss}). With the warmup strategy, the ratio of hard samples was decreased; in other words, loss function less fluctuated. As a result, we achieved the highest top-1 accuracy of 74.38\% (averaged out multiple runs) regardless of a high $\gamma=2$ value.

\begin{table}
    \centering
    \caption{Ablation studies on CIFAR-100 (ResNet-110)}\label{tab:abs_cifar}\vspace{0.2cm}
    \scalebox{0.8}{%
    \begin{tabular}{c c@{\hskip1cm} c c }
    \toprule
    & & Top-1 & Top-5 \\
    \midrule
    \multicolumn{4}{l}{\bf{Static anchor probabilities}}\\
	$\gamma=0.5$ & $\pred{*}=0.8$ & 73.74 & 92.45 \\
	$\gamma=0.5$ & $\pred{*}=0.5$ & 73.77 & 92.30 \\
	$\gamma=0.5$ & $\pred{*}=0.1$ & 73.11 & 92.08 \\
    \midrule	
    \multicolumn{4}{l}{\bf{Dynamic anchor probabilities}}\\
    $\gamma=0.5$ & - &\bf{74.25} & \bf{92.62} \\
    $\gamma=1.0$ & - & 73.59 & 92.04 \\
    $\gamma=2.0$ & - & 71.86 & 91.46 \\
    \bottomrule
    \end{tabular}}
\end{table}




\subsection{Human Pose Estimation}
We evaluate our method on two different human pose estimation datasets: single-person pose on MPII~\cite{andriluka14cvpr} and LSP~\cite{Johnson10} dataset. The single-person pose estimation problem assumes that the position and the scale information of a target person are given. 

\paragraph*{Implementation details. } For the task of human pose estimation, we use the Hourglass network~\cite{Newell2016StackedHN} as a baseline and only replace the loss function with the proposed loss during training. Note that we put sigmoid activation layer on top of the standard architecture to perform classification. Pose models are trained using Torch~\cite{torch} framework. The input size is set to 256$\times$256, batch size is 6, and the model is trained with a single NVIDIA Tesla V100 GPU. Learning rate is set to 0.001 for the first 100 epochs and dropped by half and 0.2 iteratively at every 20 epoch. Testing is held by averaging the heatmaps over six-scale image pyramid with flipping. 

\paragraph*{Datasets.} The MPII human pose dataset consists of 20k training images over 40k people performing various activities. We follow the previous training/validation split from~\cite{Tompson_CVPR15}, where 3k images from training set are used for validation. The LSP dataset~\cite{Johnson10} is composed of 11k training images with LSP extended dataset~\cite{Johnson11}, and containing mostly sports activities. 


\paragraph*{Results.} We evaluate the single-person pose estimation results on standard Percentage of Correct Keypoints (PCK) metric, which defines correct prediction if the distance between the output and the ground truth position lies in $\alpha$ with respect to the scale of the person. $\alpha$ is set to 0.5 and 0.2 in MPII and LSP dataset, respectively. PCK score for each dataset is reported in Table~\ref{tab:pck_mpii} and~\ref{tab:pck_lsp}.

For comparison, we split the performance table by hourglass-based architecture. The bottom rows are comparison between the methods built on top of Hourglass network. We achieve comparable results to the models built on top of hourglass network with more computational complexity on both datasets. We also report the validation score of the baseline method trained with mean squared error by conducting a single scale test for direct comparison between the losses in Table~\ref{tab:pose_val}. We found consistent improvements over the symmetric parts; Due to appearance similarity on the symmetric body parts, our loss function automatically penalizes more on those parts during training, without having any additional constraint for the symmetric parts.



\begin{figure}[t]
    \centering\vspace{-0.2cm}
    \includegraphics[width=1.0\linewidth]{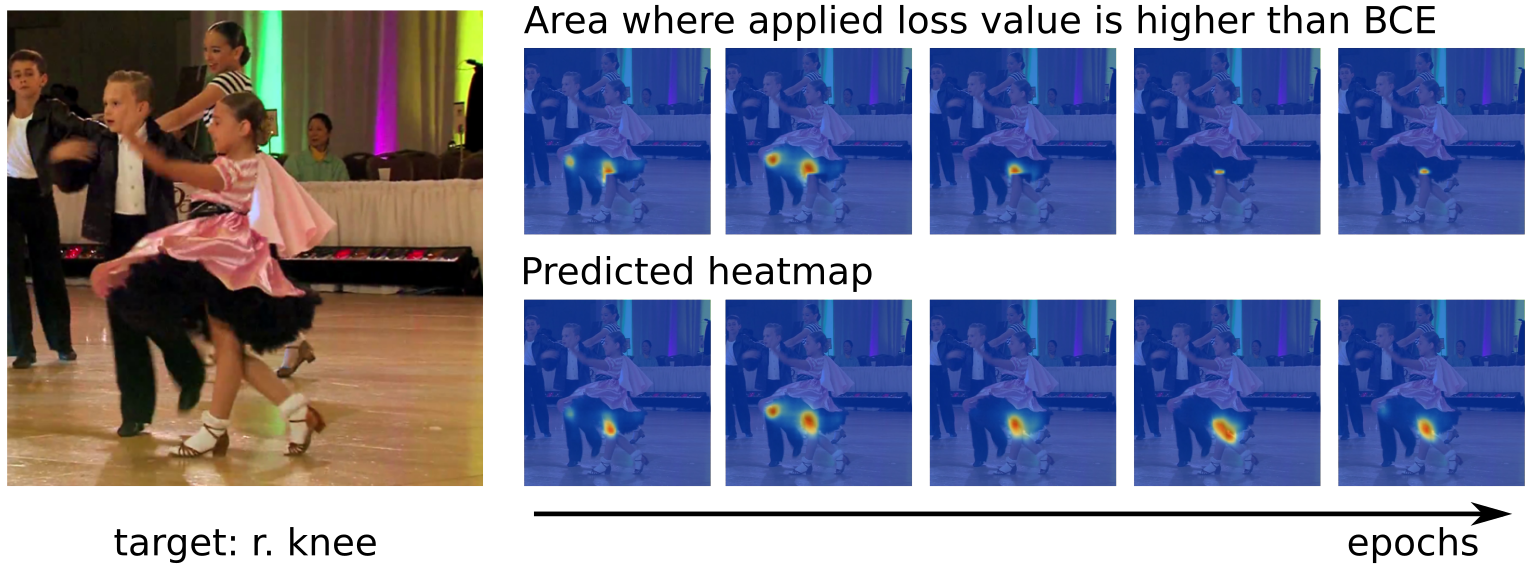}
    \caption{We visualize where anchor loss assigns higher loss values than the binary cross entropy and how it changes over training epochs. At the beginning, visually similar parts often get higher scores than the target body part, thus our loss function assigns higher weights on those pixel locations. Once the model is able to detect the target body part with high confidence, loss is down-weighted for most of the areas, so that the network can focus on finding more accurate location for the target body part.}
    \label{fig:pose}
\end{figure}

\begin{figure}[t]
    \centering\vspace{-0.2cm}
    \includegraphics[width=1.0\linewidth]{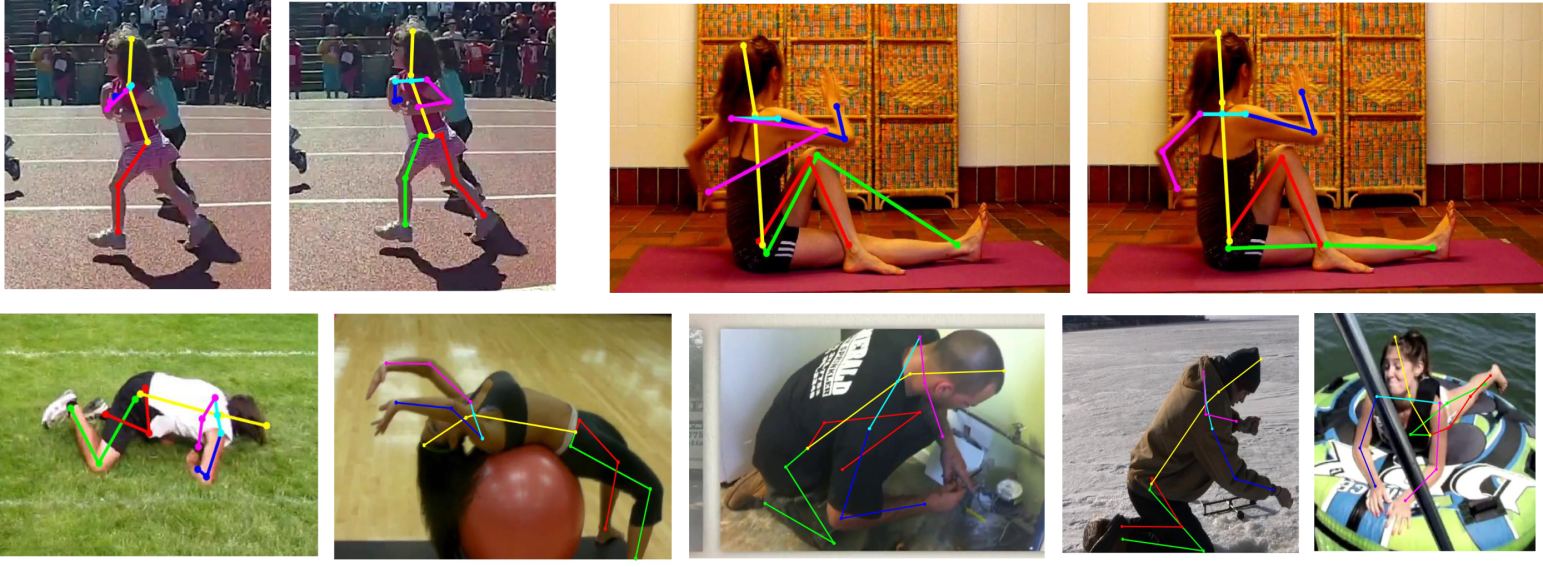}
    \caption{Qualitative results on human pose. The first row compares with the result from MSE loss (left) and our loss (right), and the second row contains some sample outputs. Model trained with the proposed loss function is robust at predicting symmetric body parts. }\label{fig:pose_samples}
\end{figure}

\begin{table}
	\caption{PCK score on MPII dataset. The bottom rows show the performances of the methods built on top of hourglass network. The model trained with anchor loss shows comparative scores to the results from more complex models. }\label{tab:pck_mpii}
	\vspace{0.2cm}
	\centering
	\resizebox{0.48\textwidth}{!}{
	\begin{tabular}{l c c c c c c c c}
	\toprule
	{\bf Method} &Head & Shoulder & Elbow & Wrist & Hip & Knee  & Ankle & Total \\
	\midrule
	Tompson~\etal~\cite{Tompson_CVPR15} & 96.1  & 91.9  & 83.9  & 77.8  & 80.9  & 72.3 & 64.8 & 82.0  \\
	Hu~\& Ramanan~\cite{Hu_CVPR16} & 95.0  & 91.6  & 83.0  & 76.6  & 81.9  & 74.5 & 69.5 & 82.4  \\
	Pishchulin~\etal~\cite{pishchulin16cvpr} & 94.1  & 90.2  & 83.4  & 77.3  & 82.6  & 75.7 & 68.6 & 82.4 \\
	Lifshitz~\etal~\cite{Lifshitz_ECCV16} & 97.8  & 93.3  & 85.7  & 80.4  & 85.3  & 76.6 & 70.2 & 85.0 \\
	Gkioxary~\etal~\cite{Gkioxari_ECCV16} & 96.2  & 93.1  & 86.7  & 82.1  & 85.2  & 81.4 & 74.1 & 86.1 \\
	Rafi~\etal~\cite{Rafi_BMVC16} & 97.2  & 93.9  & 86.4  & 81.3  & 86.8  & 80.6 & 73.4 & 86.3 \\
	Insafutdinov~\etal~\cite{insafutdinov_eccv16}& 96.8  & 95.2  & 89.3  & 84.4  & 88.4  & 83.4 & 78.0 & 88.5 \\
	Belagiannis~\& Zisserman~\cite{Belagiannis17}& 97.7  & 95.0  & 88.2  & 83.0  & 87.9  & 82.6 & 78.4 & 88.1 \\
	Wei~\etal~\cite{Wei_CVPR16}& 97.8  & 95.0  & 88.7  & 84.0  & 88.4  & 82.8 & 79.4 & 88.5  \\
	Bulat~\& Tzimiropoulos~\cite{DBLP:conf/eccv/BulatT16}& 97.9  & 95.1  & 89.9  & 85.3  & 89.4  & 85.7 & 81.7 & 89.7 \\
	Ning~\etal~\cite{Ning_TMM18}& 98.1  & 96.3  & 92.2  & 87.8  & 90.6  & 87.6 & 82.7 & 91.2  \\
	Tang~\etal~\cite{Tang_2018_ECCV} & 98.4  & \bf{96.9}  & 92.6  & \bf{88.7}  & \bf{91.8}  & 89.4 & 86.2 & \bf{92.3}  \\
    \hline
    \multicolumn{9}{l}{{\bf{Hourglass model variants}}}\\
    Chu~\etal~\cite{ChuYOMYW17}& 98.5  & 96.3  & 91.9  & 88.1  & 90.6  & 88.0 & 85.0 & 91.5  \\
    Chen~\etal~\cite{Chen17_pose}& 98.1  & 96.5  & 92.5  & 88.5  & 90.2  & \bf{89.6} & 86.0 & 91.9  \\
    Yang~\etal~\cite{YangLOLW17} & 98.5  & 96.7  & 92.5  & \bf{88.7}  & 91.1  & 88.6 & 86.0 & 92.0 \\
    Ke~\etal~\cite{KeECCV18} & 98.5  & 96.8  & \bf{92.7}  & 88.4  & 90.6  & 89.3 & \bf{86.3} & 92.1  \\
    Hourglass + MSE~\cite{Newell2016StackedHN}& 98.2  & 96.3  & 91.2  & 87.1  & 90.1  & 87.4 & 83.6 & 90.9 \\
	Hourglass + AL (Ours) & \bf{98.6}  & 96.6  & 92.3  & 87.8  & 90.8  & 88.8 & 86.0 & 91.9 \\
	\bottomrule
	\end{tabular}}
\end{table}
\begin{table}
	\caption{PCK score on LSP dataset. The bottom rows show the performances of the methods built on top of hourglass network. We achieve better performance on LSP dataset without adding the complexity, by training the network with anchor loss. For comparison, we also report the state-of-the-art score on the top row. }\label{tab:pck_lsp}
	\vspace{0.2cm}
	\centering
	\resizebox{0.48\textwidth}{!}{
	\begin{tabular}{l c c c c c c c c}
	\toprule
    {\bf Method} &Head & Shoulder & Elbow & Wrist & Hip & Knee  & Ankle & Total \\
	\midrule
    Lifshitz~\etal~\cite{Lifshitz_ECCV16} & 96.8  & 89.0  & 82.7  & 79.1  & 90.9  & 86.0 & 82.5 & 86.7 \\
    Pishchulin~\etal~\cite{pishchulin16cvpr} & 97.0  & 91.0  & 83.8  & 78.1  & 91.0  & 86.7 & 82.0 & 87.1 \\
    Insafutdinov~\etal~\cite{insafutdinov_eccv16}& 97.4  & 92.7  & 87.5  & 84.4  & 91.5  & 89.9 & 87.2 & 90.1  \\
    Wei~\etal~\cite{Wei_CVPR16} & 97.8  & 92.5  & 87.0  & 83.9  & 91.5  & 90.8 & 89.9 & 90.5 \\
    Bulat\&Tzimiropoulos~\cite{DBLP:conf/eccv/BulatT16} & 97.2  & 92.1  & 88.1  & 85.2  & 92.2  & 91.4 & 88.7 & 90.7 \\
    Ning~\etal~\cite{Ning_TMM18} & 98.2  & 94.4  & 91.8  & 89.3  & 94.7  & 95.0 & 93.5 & 93.9 \\
    Tang~\etal~\cite{Tang_2018_ECCV} & 98.3 & \bf{95.9} & \bf{93.5} & \bf{90.7} & \bf{95.0} & \bf{96.6} & \bf{95.7} & \bf{95.1} \\
    \hline
    \multicolumn{9}{l}{{\bf{Hourglass model variants}}}\\
    Chu~\etal~\cite{ChuYOMYW17}& 98.1  & 93.7  & 89.3  & 86.9  & 93.4  & 94.0 & 92.5 & 92.6 \\
    Yang~\etal~\cite{YangLOLW17} & 98.3  & 94.5  & 92.2  & 88.9  & 94.4  & 95.0 & 93.7 & 93.9 \\
    Hourglass + AL (Ours)& \bf{98.6}  & 94.8  & 92.5  & 89.3  & 93.9  & 94.8 & 94.0 & 94.0 \\
	\bottomrule
	\end{tabular}}
\end{table}

\begin{table}
	\caption{Validation Results on MPII dataset. We report the validation score of the result using different losses with the same single-scale testing setup. }\label{tab:pose_val}
	\vspace{0.2cm}
	\centering
	\resizebox{0.48\textwidth}{!}{
	\begin{tabular}{ l c c c c c c c c}
	\toprule
	{\bf Method} & Head & Shoulder & Elbow & Wrist & Hip & Knee & Ankle & Mean \\
	\midrule
	Hourglass + MSE & \bf{96.73} & 95.94 & 90.39 & 85.40 & 89.04 & 85.17 & 81.86 & 89.32 \\
	Hourglass + AL (Ours) & 96.45 & \bf{96.04} & \bf{90.46} & \bf{86.00} & \bf{89.20} & \bf{86.84} & \bf{83.68} & \bf{89.93} \\
	\bottomrule
	\end{tabular}}
\end{table}

\paragraph*{Ablation Studies. }
We conduct ablation studies by varying $\gamma$ on 2-stacked hourglass network and report the score in Table~\ref{tab:hg2_abl}. With proper selection of $\gamma=2.0$, we can achieve better performance over all the losses.

\begin{table}
	\caption{Hyperparameter search and comparison to other losses on MPII dataset with 2-stacked hourglass network.}\label{tab:hg2_abl}
	\vspace{0.2cm}
	\centering
	\resizebox{0.48\textwidth}{!}{
	\begin{tabular}{ l c c c c c c c c}
	\toprule
	\bf{Method} & Head & Shoulder & Elbow & Wrist & Hip & Knee & Ankle & Mean \\
	\midrule
	BCE & 96.42 & 95.35 & 89.82 & 84.72 & 88.47 & 85.17 & 81.13 & 88.84 \\
	MSE & 96.42 & 95.30 & 89.57 & 84.63 & 88.78 & 85.07 & \bf{81.77} & 88.89 \\
	FL & \bf{96.52} & \bf{95.47} & 89.71 & 84.87 & 88.38 & 84.75 & 81.25 & 88.81 \\
	\midrule
	AL, $\gamma$ = 5 & 96.35 & 95.04 & 89.26 & 84.56 & \bf{88.99} & \bf{85.51} & 81.37 & 88.84 \\
	AL, $\gamma$ = 1 & 96.35 & 95.40 & 89.60 & 85.11 & 88.59 & 84.85 & \bf{81.77} & 88.94 \\
	AL, $\gamma$ = 2 & 96.49 & 95.45 & \bf{90.08} & \bf{85.42} & 88.64 & 85.31 & 81.60 & \bf{89.11} \\
	\bottomrule
	\end{tabular}}
\end{table}

\paragraph*{Qualitative Analysis.} We visualize which area gets more penalty than the standard binary cross entropy in Fig~\ref{fig:pose}. For the fist few epochs, we can see that visually similar parts of both target and non-target person get higher penalty. Once the model finds the correct body part locations, the loss function is down-weighted and the area of higher penalty is focused only on few pixel locations, which helps fine adjustments on finding more accurate locations. We also show some sample outputs in Fig~\ref{fig:pose_samples}. For comparison, the top row shows some outputs from the model trained with MSE (left) and anchor loss (right). We can see that the network trained with proposed loss is robust at predicting symmetric parts. 

\paragraph*{Double-counting.} 
\begin{wrapfigure}[7]{l}[0pt]{3.2cm}
     \includegraphics[width=3.6cm]{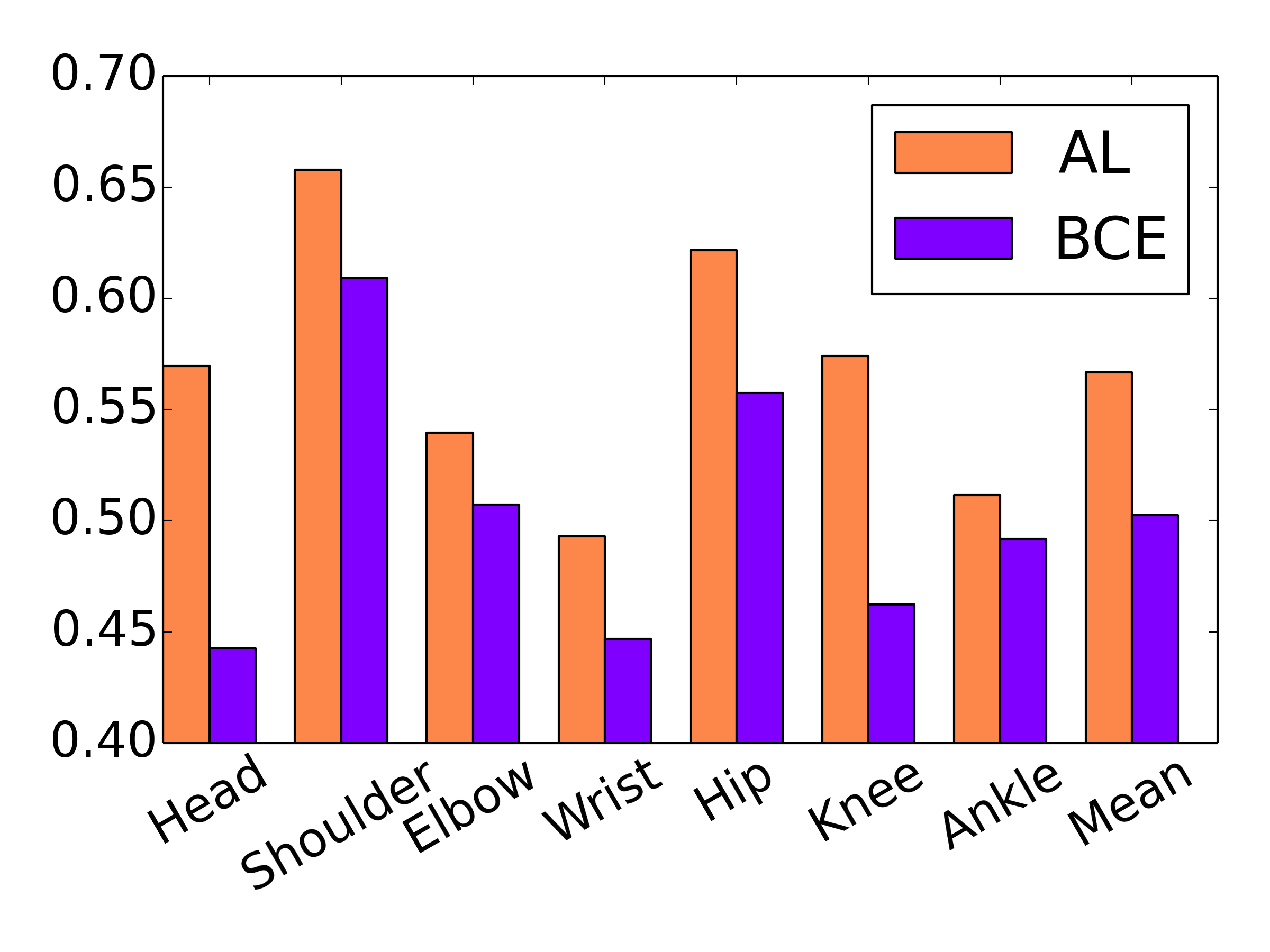}
\end{wrapfigure}
For the task of human pose estimation, we observe a double-counting problem, where the predicted heatmap shows multiple peaks. To analyze how AL behaves in those cases, we depict the ratio of the correct prediction when double-counting problems are encountered on MPII dataset. Overall, AL assigns correct body parts compared to BCE.
\section{Conclusion}
In this paper, we presented anchor loss function which adaptively rescales the standard cross entropy function based on prediction difficulty. The network automatically evaluates the prediction difficulty by measuring the divergence among the network outputs regarding true positive and false positive predictions. The proposed loss function has shown strong empirical results on two different domains: image classification and human pose estimation. A simple drop-in replacement for standard cross entropy loss gives performance improvement. With a proper selection of designing the re-weighing scheme and anchor probability, the anchor loss can be applied to diverse machine learning and computer vision applications. 

\paragraph*{Acknowledgement} We would like to thank Joseph Marino and Matteo Ruggero Ronchi for their valuable comments. This work was supported by funding from Disney Research. 

{\small
\bibliographystyle{ieee_fullname}
\bibliography{refs}

\begin{thebibliography}{10}\itemsep=-1pt

\bibitem{andriluka14cvpr}
Mykhaylo Andriluka, Leonid Pishchulin, Peter Gehler, and Bernt Schiele.
\newblock {2D} human pose estimation: New benchmark and state of the art
  analysis.
\newblock In {\em Proc. IEEE CVPR}, 2014.

\bibitem{BarronCVPR2019}
Jonathan~T. Barron.
\newblock A general and adaptive robust loss function.
\newblock In {\em Proc. IEEE CVPR}, 2019.

\bibitem{Belagiannis15}
Vasileios Belagiannis, Christian Rupprecht, Gustavo Carneiro, and Nassir Navab.
\newblock Robust optimization for deep regression.
\newblock In {\em Proc. IEEE ICCV}, 2015.

\bibitem{Belagiannis17}
Vasileios Belagiannis and Andrew Zisserman.
\newblock Recurrent human pose estimation.
\newblock In {\em Proc. IEEE FG}, 2017.

\bibitem{Berrada18}
Leonard Berrada, Andrew Zisserman, and M.~Pawan Kumar.
\newblock Smooth loss functions for deep top-k classification.
\newblock In {\em ICLR}, 2018.

\bibitem{NN_BUDA2018}
Mateusz Buda, Atsuto Maki, and Maciej~A. Mazurowski.
\newblock A systematic study of the class imbalance problem in convolutional
  neural networks.
\newblock {\em Neural Networks}, 106:249 -- 259, 2018.

\bibitem{DBLP:conf/eccv/BulatT16}
Adrian Bulat and Georgios Tzimiropoulos.
\newblock Human pose estimation via convolutional part heatmap regression.
\newblock In {\em ECCV}, 2016.

\bibitem{Chawla02smote}
Nitesh~V. Chawla, Kevin~W. Bowyer, Lawrence~O. Hall, and W.~Philip Kegelmeyer.
\newblock Smote: Synthetic minority over-sampling technique.
\newblock {\em Journal of Artificial Intelligence Research}, 16:321--357, 2002.

\bibitem{Chen17_pose}
Yu Chen, Chunhua Shen, Xiu-Shen Wei, Lingqiao Liu, and Jian Yang.
\newblock Adversarial posenet: {A} structure-aware convolutional network for
  human pose estimation.
\newblock In {\em Proc. IEEE ICCV}, 2017.

\bibitem{CPN17}
Yilun Chen, Zhicheng Wang, Yuxiang Peng, Zhiqiang Zhang, Gang Yu, and Jian Sun.
\newblock Cascaded pyramid network for multi-person pose estimation.
\newblock In {\em Proc. IEEE CVPR}, 2017.

\bibitem{ChuYOMYW17}
Xiao Chu, Wei Yang, Wanli Ouyang, Cheng Ma, Alan~L. Yuille, and Xiaogang Wang.
\newblock Multi-context attention for human pose estimation.
\newblock In {\em {Proc. IEEE CVPR}}, 2017.

\bibitem{torch}
R. Collobert, K. Kavukcuoglu, and C. Farabet.
\newblock Torch7: A matlab-like environment for machine learning.
\newblock In {\em BigLearn, NIPS Workshop}, 2011.

\bibitem{imagenet_cvpr09}
Jia Deng, Wei Dong, Richard Socher, Li jia Li, Kai Li, and Fei-Fei Li.
\newblock {ImageNet}: A large-scale hierarchical image database.
\newblock In {\em {Proc. IEEE CVPR}}, 2009.

\bibitem{ICCV17_Dong}
Qi Dong, Shaogang Gong, and Xiatian Zhu.
\newblock Class rectification hard mining for imbalanced deep learning.
\newblock In {\em Proc. IEEE ICCV}, 2017.

\bibitem{FelzenszwalbGM10}
Pedro~F. Felzenszwalb, Ross~B. Girshick, and David~A. McAllester.
\newblock Cascade object detection with deformable part models.
\newblock In {\em {Proc. IEEE CVPR}}, 2010.

\bibitem{Gkioxari_ECCV16}
Georgia Gkioxari, Alexander Toshev, and Navdeep Jaitly.
\newblock Chained predictions using convolutional neural networks.
\newblock In {\em ECCV}, 2016.

\bibitem{Gong14WARP}
Yunchao Gong, Yangqing Jia, Thomas~K. Leung, Alexander Toshev, and Sergey
  Ioffe.
\newblock deep convolutional ranking for multi label image annotation.
\newblock In {\em ICLR}, 2014.

\bibitem{Han05}
Hui Han, Wen-Yuan Wang, and Bing-Huan Mao.
\newblock {Borderline-SMOTE}: A new over-sampling method in imbalanced data
  sets learning.
\newblock In {\em ICIC}, 2005.

\bibitem{He2016DeepRL}
Kaiming He, Xiangyu Zhang, Shaoqing Ren, and Jian Sun.
\newblock Deep residual learning for image recognition.
\newblock In {\em Proc. IEEE CVPR}, 2016.

\bibitem{NIPS2018_Hendrycks}
Dan Hendrycks, Mantas Mazeika, Duncan Wilson, and Kevin Gimpel.
\newblock Using trusted data to train deep networks on labels corrupted by
  severe noise.
\newblock In {\em NuerIPS}, 2018.

\bibitem{Hoiem_ECCV12}
Derek Hoiem, Yodsawalai Chodpathumwan, and Qieyun Dai.
\newblock Diagnosing error in object detectors.
\newblock In {\em ECCV}, 2012.

\bibitem{Hu_CVPR16}
Peiyun Hu and Deva Ramanan.
\newblock Bottom-up and top-down reasoning with hierarchical rectified
  gaussians.
\newblock In {\em Proc. IEEE CVPR}, 2016.

\bibitem{CVPR_HuangLLT16}
Chen Huang, Yining Li, Chen~Change Loy, and Xiaoou Tang.
\newblock Learning deep representation for imbalanced classification.
\newblock In {\em Proc. IEEE CVPR}, 2016.

\bibitem{huber_1964}
Peter~J. Huber.
\newblock Robust estimation of a location parameter.
\newblock {\em Annals of Mathematical Statistics}, 35(1):73--101, Mar. 1964.

\bibitem{insafutdinov_eccv16}
Eldar Insafutdinov, Leonid Pishchulin, Bjoern Andres, Mykhaylo Andriluka, and
  Bernt Schiele.
\newblock {DeeperCut}: A deeper, stronger, and faster multi-person pose
  estimation model.
\newblock In {\em ECCV}, 2016.

\bibitem{Johnson10}
Sam Johnson and Mark Everingham.
\newblock Clustered pose and nonlinear appearance models for human pose
  estimation.
\newblock In {\em BMVC}, 2010.

\bibitem{Johnson11}
Sam Johnson and Mark Everingham.
\newblock Learning effective human pose estimation from inaccurate annotation.
\newblock In {\em Proc. IEEE CVPR}, 2011.

\bibitem{KeECCV18}
Lipeng Ke, Ming-Ching Chang, Honggang Qi, and Siwei Lyu.
\newblock Multi-scale structure-aware network for human pose estimation.
\newblock In {\em ECCV}, 2018.

\bibitem{KrizhevskyCIFAR}
Alex Krizhevsky.
\newblock Learning multiple layers of features from tiny images.
\newblock Technical report, 2009.

\bibitem{Lifshitz_ECCV16}
Ita Lifshitz, Ethan Fetaya, and Shimon Ullman.
\newblock Human pose estimation using deep consensus voting.
\newblock In {\em ECCV}, 2016.

\bibitem{FocalLoss17}
Tsung{-}Yi Lin, Priya Goyal, Ross~B. Girshick, Kaiming He, and Piotr
  Doll{\'{a}}r.
\newblock Focal loss for dense object detection.
\newblock In {\em {Proc. IEEE ICCV}}, 2017.

\bibitem{SSD15}
Wei Liu, Dragomir Anguelov, Dumitru Erhan, Christian Szegedy, Scott Reed,
  Cheng-Yang Fu, and Alexander~C. Berg.
\newblock {SSD}: Single shot multibox detector.
\newblock In {\em ECCV}, 2015.

\bibitem{LiuWYY16}
Weiyang Liu, Yandong Wen, Zhiding Yu, and Meng Yang.
\newblock Large-margin softmax loss for convolutional neural networks.
\newblock In {\em ICML}, 2016.

\bibitem{Newell2016StackedHN}
Alejandro Newell, Kaiyu Yang, and Jia Deng.
\newblock Stacked hourglass networks for human pose estimation.
\newblock In {\em ECCV}, 2016.

\bibitem{Ning_TMM18}
Guanghan Ning, Zhi Zhang, and Zhiquan He.
\newblock Knowledge-guided deep fractal neural networks for human pose
  estimation.
\newblock {\em {IEEE} Trans. Multimedia}, 20(5):1246--1259, 2018.

\bibitem{paszke2017automatic}
Adam Paszke, Sam Gross, Soumith Chintala, Gregory Chanan, Edward Yang, Zachary
  DeVito, Zeming Lin, Alban Desmaison, Luca Antiga, and Adam Lerer.
\newblock Automatic differentiation in pytorch.
\newblock In {\em NIPS-Workshops}, 2017.

\bibitem{pishchulin16cvpr}
Leonid Pishchulin, Eldar Insafutdinov, Siyu Tang, Bjoern Andres, Mykhaylo
  Andriluka, Peter Gehler, and Bernt Schiele.
\newblock {DeepCut}: Joint subset partition and labeling for multi person pose
  estimation.
\newblock In {\em Proc. IEEE CVPR}, 2016.

\bibitem{ICML18_Ren}
Mengye Ren, Wenyuan Zeng, Bin Yang, and Raquel Urtasun.
\newblock Learning to reweight examples for robust deep learning.
\newblock In {\em ICML}, 2018.

\bibitem{Ronchi_2017_ICCV}
Matteo~Ruggero Ronchi and Pietro Perona.
\newblock Benchmarking and error diagnosis in multi-instance pose estimation.
\newblock In {\em Proc. IEEE ICCV}, 2017.

\bibitem{ShrivastavaGG16}
Abhinav Shrivastava, Abhinav Gupta, and Ross~B. Girshick.
\newblock Training region-based object detectors with online hard example
  mining.
\newblock In {\em Proc. IEEE CVPR}, 2016.

\bibitem{Sung96}
Kah~Kay Sung.
\newblock {\em Learning and Example Selection for Object and Pattern
  Detection}.
\newblock PhD thesis, 1996.

\bibitem{Tang_2018_ECCV}
Wei Tang, Pei Yu, and Ying Wu.
\newblock Deeply learned compositional models for human pose estimation.
\newblock In {\em ECCV}, 2018.

\bibitem{Tompson_CVPR15}
Jonathan Tompson, Ross Goroshin, Arjun Jain, Yann LeCun, and Christoph Bregler.
\newblock Efficient object localization using convolutional networks.
\newblock In {\em Proc. IEEE CVPR}, 2015.

\bibitem{Rafi_BMVC16}
Juergen~Gall Umer~Rafi, Bastian~Leibe and Ilya Kostrikov.
\newblock An efficient convolutional network for human pose estimation.
\newblock In {\em BMVC}, 2016.

\bibitem{Wei_CVPR16}
Shih-En Wei, Varun Ramakrishna, Takeo Kanade, and Yaser Sheikh.
\newblock Convolutional pose machines.
\newblock In {\em Proc. IEEE CVPR}, 2016.

\bibitem{YangLOLW17}
Wei Yang, Shuang Li, Wanli Ouyang, Hongsheng Li, and Xiaogang Wang.
\newblock Learning feature pyramids for human pose estimation.
\newblock In {\em {Proc. IEEE ICCV}}, 2017.

\bibitem{Zhang06_mll}
Min-Ling Zhang and Zhi-Hua Zhou.
\newblock Multilabel neural networks with applications to functional genomics
  and text categorization.
\newblock {\em IEEE Trans.\ Knowl.\ Data Eng.}, 18(10):1338--1351, Oct. 2006.

\bibitem{Zhang_ICML04}
Tong Zhang.
\newblock Solving large scale linear prediction problems using stochastic
  gradient descent algorithms.
\newblock In {\em ICML}, 2004.

\bibitem{NIPS2018_Zhang}
Zhilu Zhang and Mert Sabuncu.
\newblock Generalized cross entropy loss for training deep neural networks with
  noisy labels.
\newblock In {\em NeurIPS}, 2018.

\bibitem{Zhou_KDE06}
Zhi-Hua Zhou and Xu-Ying Liu.
\newblock Training cost-sensitive neural networks with methods addressing the
  class imbalance problem.
\newblock {\em IEEE Trans.\ Knowl.\ Data Eng.}, 18(1):63--77, Feb. 2006.

\end{thebibliography}
}

\setcounter{section}{0}
\setcounter{equation}{0}
\setcounter{figure}{0}
\setcounter{table}{0}
\renewcommand{\theequation}{A-\arabic{equation}}
\renewcommand{\thefigure}{A-\arabic{figure}}
\renewcommand{\thetable}{A-\arabic{figure}}
\renewcommand\thesubsection{A-\arabic{subsection}}

\section*{Appendix}
\subsection{Anchor design}

In the paper, we set the anchor probability to the target class prediction score and modulate loss of the background class.
Here we further study how to design anchor probability that affects behavior of the loss. We first define the basic formulation of anchor loss (AL) with sigmoid-binary cross entropy:
\begin{align}
    \lossfunc{} (p,\pred{};\gamma) &= - \underbrace{(1 - \pred{} + \pred{pos})^{\gamma_t} p \log(\pred{})}_{\textrm{target class}} \\\nonumber
    &-\underbrace{(1+\pred{}-\pred{neg})^{\gamma_b} (1-p)\log(1-\pred{})}_{\textrm{background class}}.
\end{align}

Anchor probability is a reference value for determining the prediction difficulty, which is defined as a confidence score gap between the target and background classes. The prediction difficulty is used to modulate loss values either by (i) pushing the loss of target class high, (ii) suppressing the loss of background classes, or (iii) using both ways around. The details of parameter setting for each case are as follows:
\begin{enumerate}[(i)]
    \item {\bf{Modulate loss for target class}}: We set the anchor probability to the maximum prediction score among background classes. Hence, target class loss gets more penalty when its score is lower than the anchor probability.
    \begin{gather}
        \pred{*} = \max_{i, \forall p_i = 0} q_i, \nonumber\\
        \text{$\gamma_{t}=\gamma$ and $\gamma_{b}=0$}.
    \end{gather}
    \item {\bf{Modulate loss for background classes}}: We set the anchor probability to prediction score of the target class. Anchor loss is penalized more when output scores of the background classes are higher than the target. 
    \begin{gather}
        \pred{neg} = \pred{j},~\text{for $j$, $p_j = 1$},\nonumber\\
        \text{$\gamma_{t}=0$ and $\gamma_{b}=\gamma$}.
    \end{gather}
    \item {\bf{Modulate loss for both target and background classes}}: We modulate loss on both directions by combining the above cases.
    \begin{gather}
        \pred{pos} = \max_{i, \forall p_i = 0} q_i, \nonumber\\
        \pred{neg} = \pred{j},~\text{for $j$, $p_j = 1$},\\
        \gamma_{t} = \gamma_{b} = \gamma.\nonumber
    \end{gather}
\end{enumerate}

\begin{figure}
\centering
\footnotesize
 \begin{center}
   \subfigure[Modulate target loss]{\includegraphics[width=6.8cm]{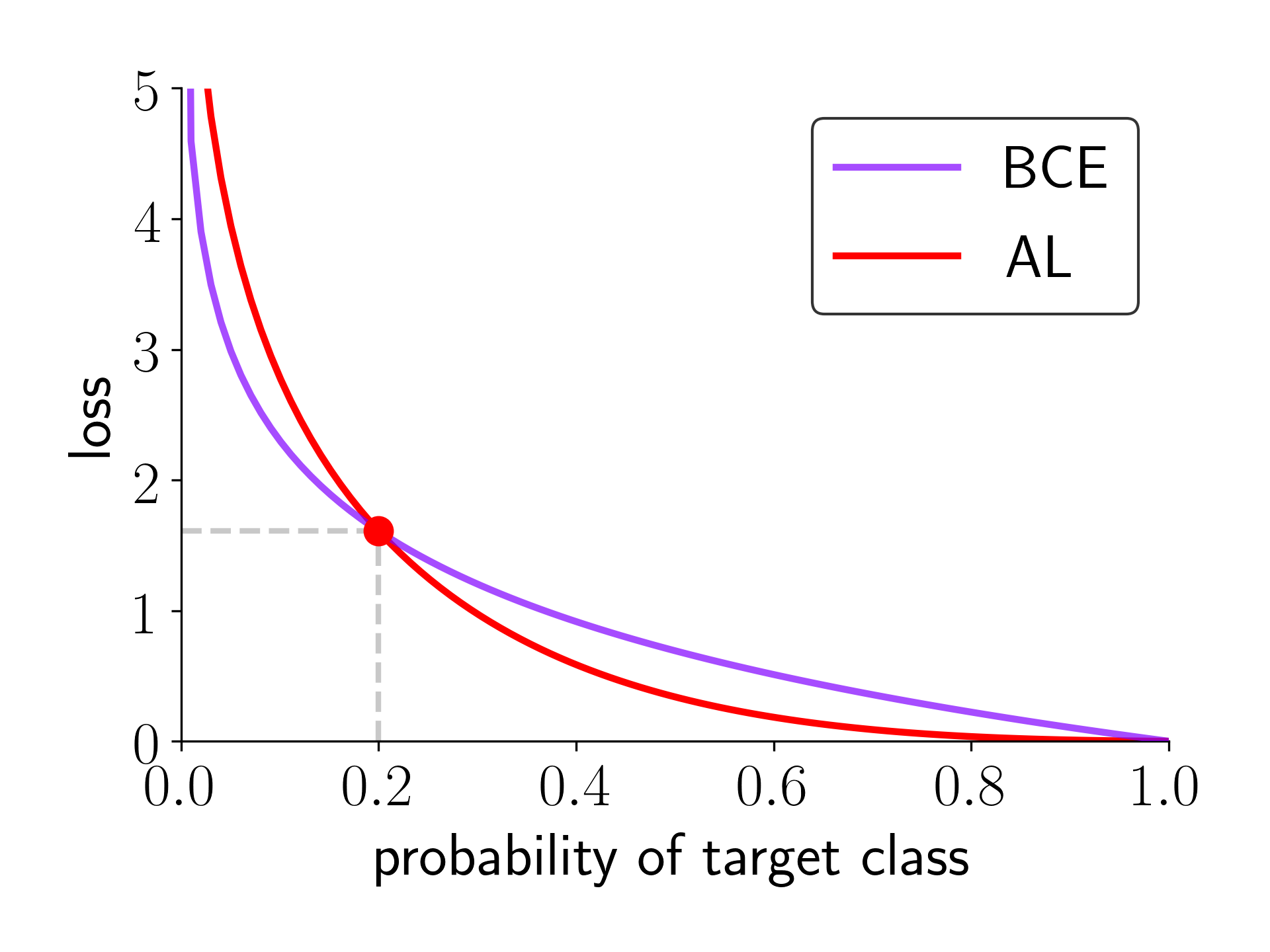}}\!
   \subfigure[Modulate background loss]{\includegraphics[width=6.8cm]{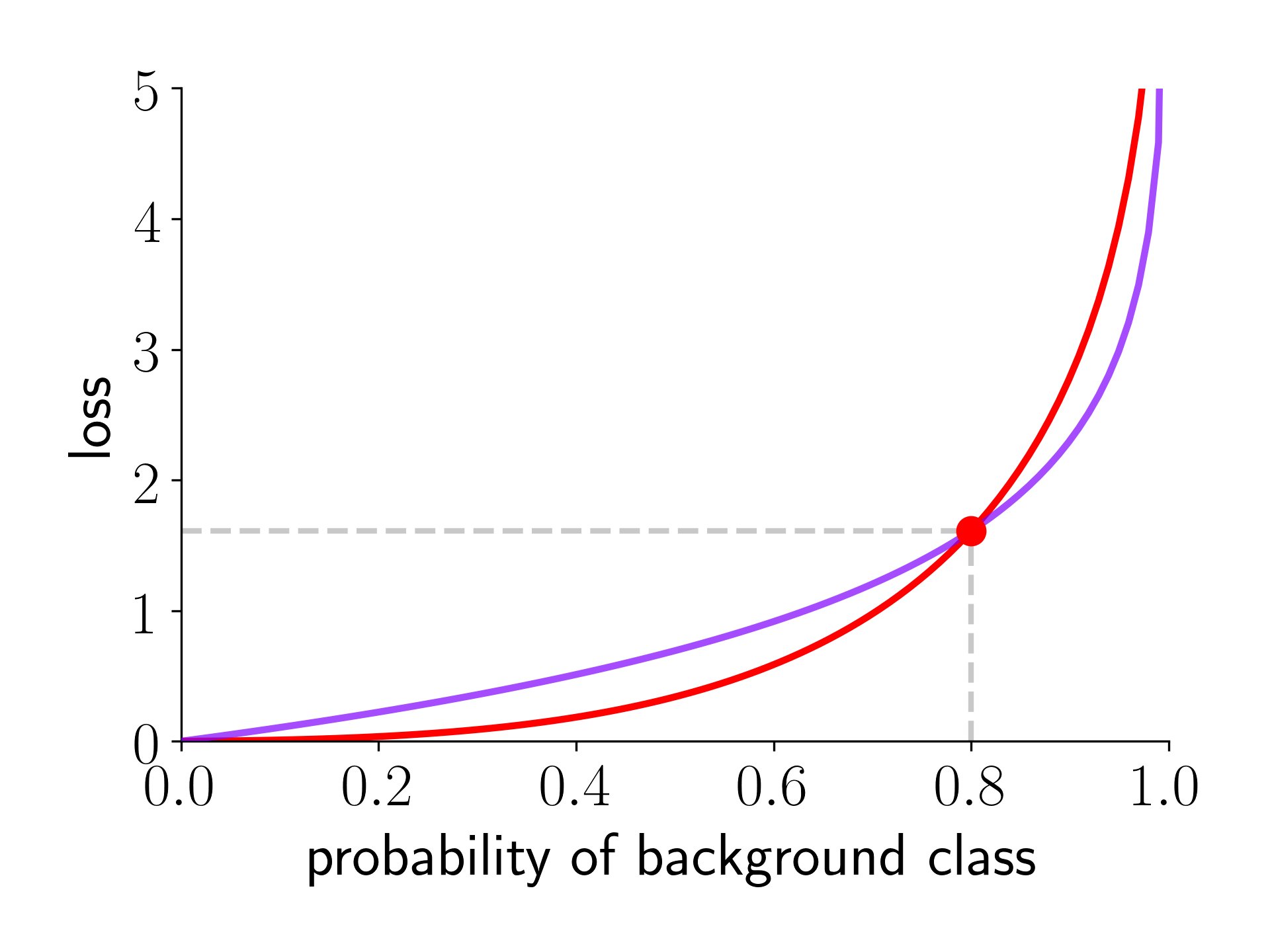}}
 \end{center}
   \caption{How an anchor probability modulates loss values. When the prediction score of target class is lower than $\pred{pos}=0.2$, anchor loss penalizes more than binary cross entropy (a). On the contrary, when the prediction score of background class is higher than $\pred{neg}=0.8$, the loss value becomes higher than the binary cross entropy (b). }
\label{fig:grad}
\end{figure}

We report image classification performance on CIFAR-100 by varying the way of designing anchor probability in Table~\ref{tab:diff_anchor}. We achieve the best performance by modulating the loss for background classes (ii).


\begin{table}[b]
	\caption{Classification accuracies on CIFAR-100 with different anchor probabilities}\label{tab:diff_anchor}
	\vspace{0.2cm}
	\centering
	\begin{tabular}{c c c }
	\toprule
	loss fn. & Top-1 & Top-5 \\
	\midrule
	BCE & 73.88 $\pm$ 0.22 & 92.03 $\pm$ 0.42 \\
	(i) & 74.06 $\pm$ 0.53 & 92.32 $\pm$ 0.24 \\
	(ii) & \bf{74.25} $\pm$ 0.34 & \bf{92.62} $\pm$ 0.50 \\
	(iii) & 73.90 $\pm$ 0.40 & 92.24 $\pm$ 0.06 \\
	\bottomrule
	\end{tabular}
\end{table}




\begin{figure*} 
    \centering
    \includegraphics[width=16cm]{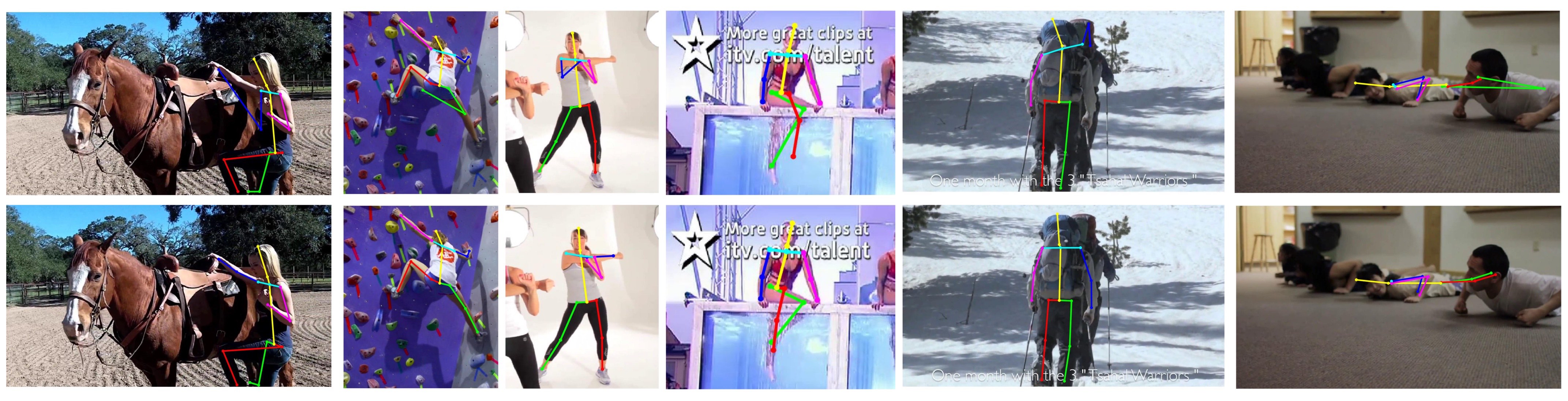}\vspace{0.2cm}
    \caption{Qualitative results for human pose estimation. Top row shows the output images with baseline (MSE) and bottom row represents the outcomes with anchor loss. }\label{fig:pose_fig}\vspace{0.5cm}
    
    \includegraphics[width=16cm]{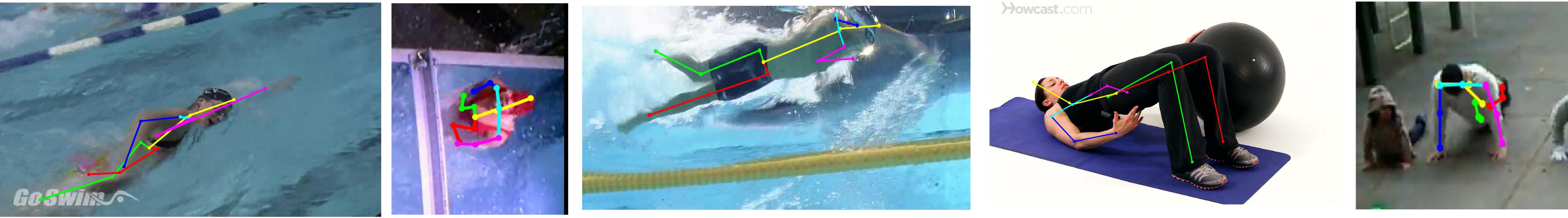}\vspace{0.2cm}
    \caption{Failure cases on human pose estimation. Network trained with anchor loss still fails to detect correct body part locations when the body part is blurred or self-occluded. }\label{fig:pose_fail}\vspace{0.5cm}
    
    \footnotesize
    \begin{tabular}{c@{\hskip1pt}c@{\hskip1pt}c@{\hskip1pt}c@{\hskip1pt}c@{\hskip1pt}c@{\hskip1pt}c@{\hskip1pt}c@{\hskip1pt}c}
    &\includegraphics[width=2cm]{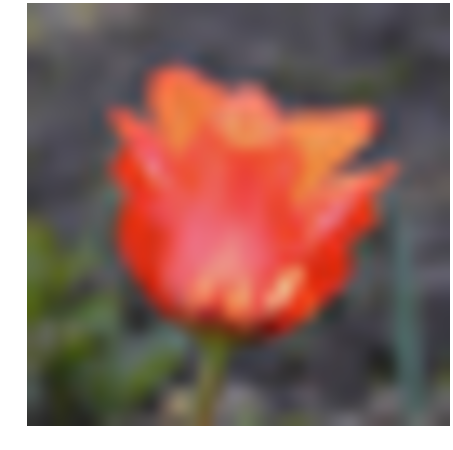}&
    \includegraphics[width=2cm]{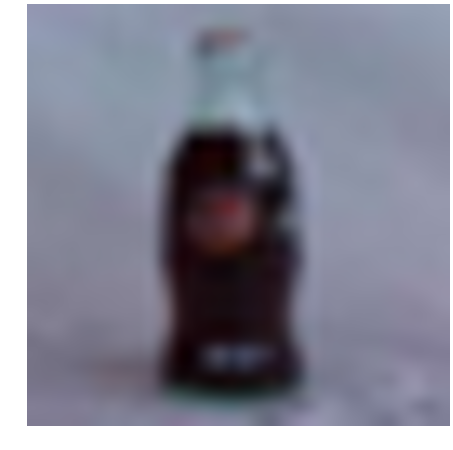}&
    \includegraphics[width=2cm]{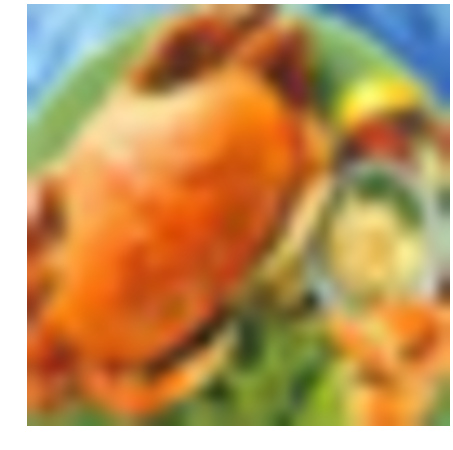}&
    \includegraphics[width=2cm]{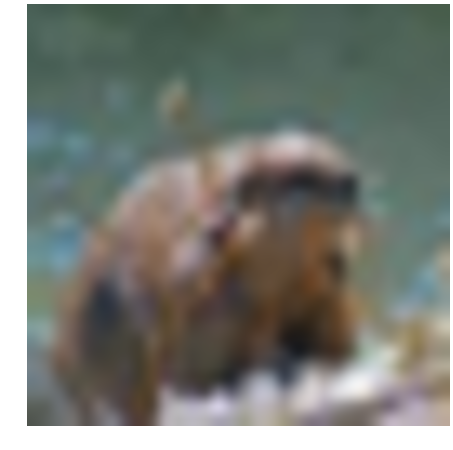}&
    \includegraphics[width=2cm]{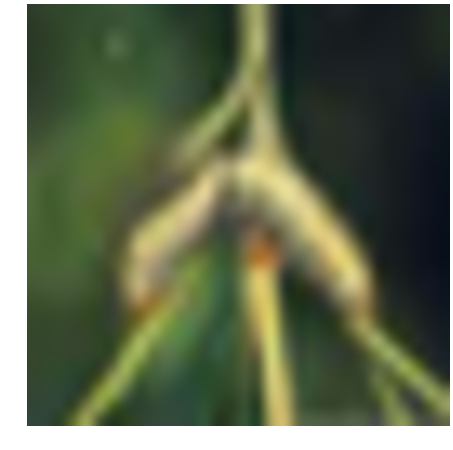}&
    \includegraphics[width=2cm]{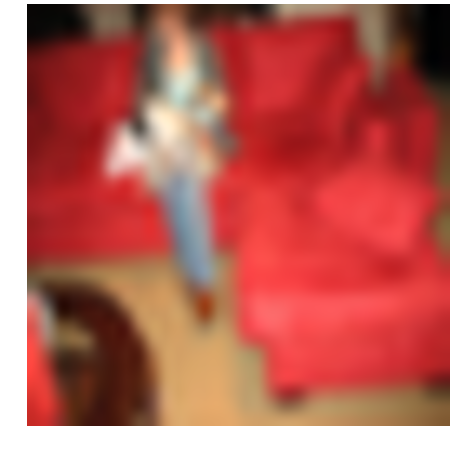}&
    \includegraphics[width=2cm]{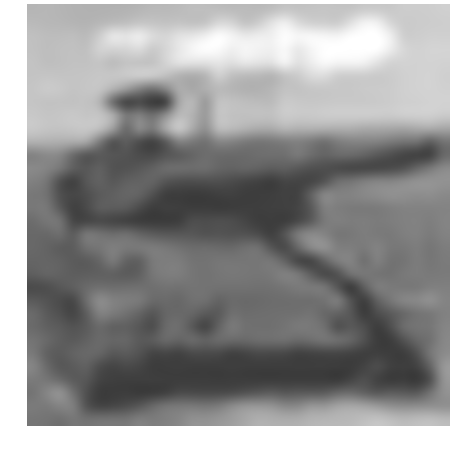}&
    \includegraphics[width=2cm]{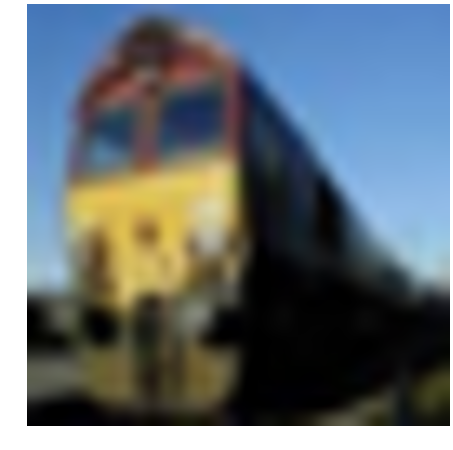}\\
    {\bf GT}&tulip&bottle&crab&beaver&sea&couch&tank&train\\
    \midrule
    {\bf CE}&\includegraphics[valign=m,width=2cm]{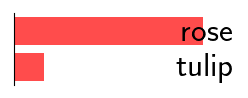}&
    \includegraphics[valign=m,width=2cm]{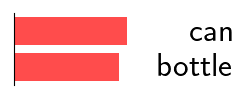}&
    \includegraphics[valign=m,width=2cm]{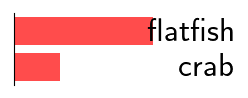}&
    \includegraphics[valign=m,width=2cm]{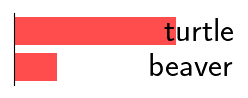}&
    \includegraphics[valign=m,width=2cm]{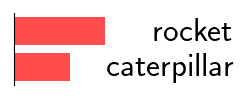}&
    \includegraphics[valign=m,width=2cm]{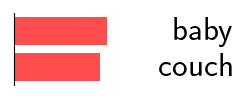}&
    \includegraphics[valign=m,width=2cm]{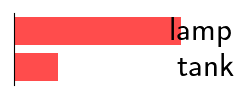}&
    \includegraphics[valign=m,width=2cm]{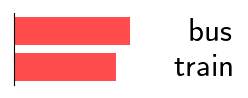}\\
    {\bf AL}&\includegraphics[valign=m,width=2cm]{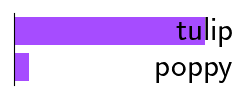}&
    \includegraphics[valign=m,width=2cm]{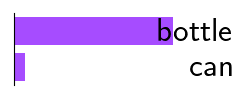}&
    \includegraphics[valign=m,width=2cm]{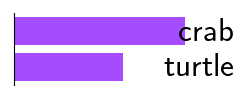}&
    \includegraphics[valign=m,width=2cm]{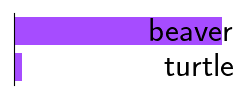}&
    \includegraphics[valign=m,width=2cm]{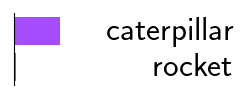}&
    \includegraphics[valign=m,width=2cm]{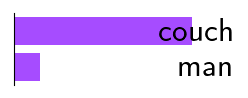}&
    \includegraphics[valign=m,width=2cm]{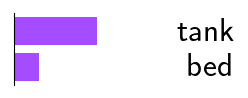}&
    \includegraphics[valign=m,width=2cm]{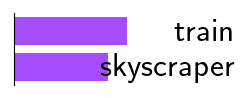}\\
    \end{tabular}\vspace{0.5cm}
	\caption{Image classification results on CIFAR-100. We compare the top-2 prediction scores of ResNet-110 with cross entropy (CE) and anchor loss (AL). Network trained with anchor loss successfully classifies difficult examples even though the model trained with cross entropy fails.}\label{fig:classif}
\end{figure*}

\subsection{Qualitative figures}
We visualize qualitative results for human pose estimation (Fig.~\ref{fig:pose_fig},~\ref{fig:pose_fail}) and image classification (Fig.~\ref{fig:classif}). Network trained with anchor loss has shown improvement over the baseline losses for both tasks. Specifically, anchor loss shows its potential use for multi-person pose estimation by finding correct body parts when the target person is occluded or overlapped by other person (last two columns of Fig.~\ref{fig:pose_fig}).




\end{document}